\documentclass[10pt,twocolumn,letterpaper]{article}

\usepackage[pagenumbers]{cvpr} 

\usepackage{graphicx}
\usepackage{amsmath}
\usepackage{amssymb}
\usepackage{booktabs}
\usepackage{xcolor}
\usepackage{bm}
\usepackage{pifont}
\usepackage{enumitem}
\usepackage{soul}

\usepackage{dblfloatfix}
\usepackage{siunitx}
\usepackage{balance}

\usepackage[pagebackref,breaklinks,colorlinks]{hyperref}

\def \ie {\emph{i.e.},}
\def \eg {\emph{e.g.},}

\newcommand{\tit}[1]{\smallbreak\noindent\textbf{#1.}}

\newcommand{\numfonts}{\num{10400} }
\newcommand{\numwords}{\num{10400} }

\usepackage[capitalize]{cleveref}
\crefname{section}{Sec.}{Secs.}
\Crefname{section}{Section}{Sections}
\Crefname{table}{Table}{Tables}
\crefname{table}{Tab.}{Tabs.}

\begin{document}

\title{Handwritten Text Generation from Visual Archetypes}

\author{Vittorio Pippi, Silvia Cascianelli, Rita Cucchiara\\
University of Modena and Reggio Emilia\\
Via Pietro Vivarelli, 10, Modena (Italy)\\
{\tt\small \{name.surname\}@unimore.it}
}
\maketitle

\begin{abstract}
Generating synthetic images of handwritten text in a writer-specific style is a challenging task, especially in the case of unseen styles and new words, and even more when these latter contain characters that are rarely encountered during training. While emulating a writer's style has been recently addressed by generative models, the generalization towards rare characters has been disregarded. In this work, we devise a Transformer-based model for Few-Shot styled handwritten text generation and focus on obtaining a robust and informative representation of both the text and the style. In particular, we propose a novel representation of the textual content as a sequence of dense vectors obtained from images of symbols written as standard GNU Unifont glyphs, which can be considered their \emph{visual archetypes}. This strategy is more suitable for generating characters that, despite having been seen rarely during training, possibly share visual details with the frequently observed ones. As for the style, we obtain a robust representation of unseen writers' calligraphy by exploiting specific pre-training on a large synthetic dataset. Quantitative and qualitative results demonstrate the effectiveness of our proposal in generating words in unseen styles and with rare characters more faithfully than existing approaches relying on independent one-hot encodings of the characters. 
\end{abstract}

\section{Introduction}
\label{sec:introduction}

\begin{figure}
    \centering
    \includegraphics[width=\columnwidth]{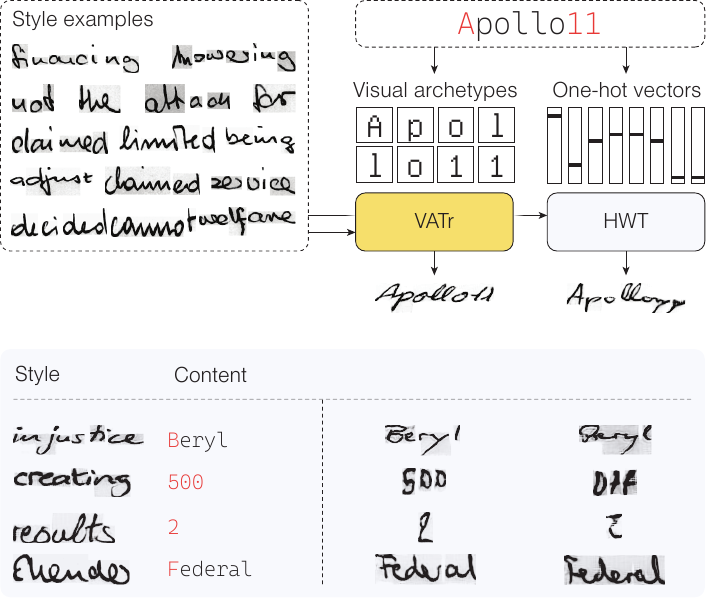} \\
    \caption{Different from previous approaches that use independent one-hot vectors as input text tokens (\eg~the State-of-the-Art HWT~\cite{bhunia2021handwriting}), we exploit \emph{visual archetypes}, \ie~geometrically-related binary images of characters. By resorting to similarities between the archetypes, we are able to generate both characters that are rarely seen during training (highlighted in red) and frequently observed ones more faithfully.\vspace{-.5cm}}
    \label{fig:first_page}
\end{figure}

Styled handwritten text generation (HTG) is an emerging research area aimed at producing writer-specific handwritten text images mimicking their calligraphic style~\cite{bhunia2021handwriting, fogel2020scrabblegan, kang2020ganwriting}. The practical applications of this research topic range from the synthesis of high-quality training data for personalized Handwritten Text Recognition (HTR) models~\cite{bhunia2021metahtr, bhunia2021text, zhang2019sequence, bhunia2019handwriting, kang2021content, kang2020distilling} to the automatic generation of handwritten notes for physically impaired people. Moreover, the writer-specific style representations that can be obtained as a by-product of models designed for this task can be applied to other tasks such as writer identification, signature verification, and handwriting style manipulation. When focusing on styled handwriting generation, simply adopting style transfer is limiting. In fact, imitating a specific writer's calligraphy does not only concern texture (\eg~the color and texture of background and ink), nor just stroke thickness, slant, skew, and roundness, but also single characters shape and ligatures. Moreover, these visual aspects must be handled properly to avoid artifacts that might result in content change (\eg~even small additional or missing strokes). 

In sight of this, specific approaches have been designed for HTG. The handwriting can be handled in the form of a trajectory (made of the underlying strokes), as done in~\cite{graves2013generating, aksan2018deepwriting, aksan2018stcn, ji2019generative, kotani2020generating}, or of an image that captures its appearance, as done in~\cite{wang2005combining, lin2007style, thomas2009synthetic, haines2016my, alonso2019adversarial, fogel2020scrabblegan, davis2020text, kang2020ganwriting, mattick2021smartpatch, gan2021higan, gan2022higan+, bhunia2021handwriting, krishnan2021textstylebrush, luo2022slogan}. The former approaches adopt online HTG strategies that entail predicting the pen trajectory point-by-point, while the latter ones are offline HTG models that output entire text images directly. We follow the offline HTG paradigm since it has the advantage, over the online one, of not requiring costly pen-recording training data, and thus, being applicable also to scenarios where the information on online handwriting is not available for a specific author (\eg~in the case of historical data) and being easier to train for not suffering of vanishing gradient and being parallelizable.

Specifically, in this work, we focus on the \emph{Few-Shot styled offline HTG} task, in which we have just a few example images of the writer's style to mimic. State-of-the-Art (SotA) approaches tackling this scenario feature an encoder that extracts writer-specific style features and a generative component, which is fed with the style features and the content representations, and produces styled text images conditioned on the desired content. These approaches usually exploit Generative Adversarial Networks (GANs~\cite{goodfellow2014generative, mirza2014conditional}), for example~\cite{alonso2019adversarial, fogel2020scrabblegan, davis2020text, kang2020ganwriting, gan2021higan, gan2022higan+, krishnan2021textstylebrush, luo2022slogan}. A more recent approach~\cite{bhunia2021handwriting} is based on an encoder-decoder generative Transformer model~\cite{vaswani2017attention} that captures character-level style variations better than previous GAN-based strategies thanks to the cross-attention mechanism between style representation and content tokens. In the approaches mentioned above, the encoding of the text content is obtained by starting from one-hot vectors, each representing a different character in a fixed charset. In this way, the characters are all independent by design. Thus, possible geometric and visual similarity among them cannot be modeled nor exploited for generation, which might result in a quality gap between the images generated by these approaches for characters that are highly represented in the training set and rare ones (\ie~long-tail characters). Moreover, for computational tractability, the fixed charset that the approaches relying on a one-hot representation of text tokens can handle is relatively small. 

\tit{Contribution}
Our proposed approach entails representing characters as \emph{continuous variables} and using them as query content vectors of a Transformer decoder for generation. In this way, the generation of characters appearing rarely in the training set (such as numbers, capital letters, and punctuation) is eased by exploiting the low distance in the latent space between rare symbols and more frequent ones (see Figure~\ref{fig:first_page}). In particular, we start from the GNU Unifont font and render each character as a $16{\times}16$ binary image, which can be considered as the \emph{visual archetype} of that character. Then, we learn a dense encoding of the character images and feed such encodings to a Transformer decoder as queries to attend the style vectors extracted by a Transformed encoder. Note that, by resorting to character images rendered in the richer GNU Unifont, which is the most complete in terms of contained Unicode characters, we can handle a huge charset (more than \num{55}k characters) seamlessly, \ie~without the need for additional parameters, as it is the case for the commonly-adopted one-hot encoding. 
Moreover, as for the style encoding part, we exploit a backbone to represent the style example images that has been pre-trained on a large synthetic dataset specifically built to focus on the calligraphic style attributes. This strategy, widely adopted for other tasks, is usually disregarded in HTG. Nonetheless, we demonstrate its effectiveness in leading to strong style representations, especially for unseen styles. We validate our proposal with extensive experimental comparison against recent generative SotA approaches, both quantitatively and qualitatively, and demonstrate the effectiveness of our proposal in generating words with both common and rare characters and in both seen and unseen styles. We call our approach VATr: Visual Archetypes-based Transformer. The code and trained models are available at \url{https://github.com/aimagelab/VATr}.

\section{Related Work}
\label{sec:related}

HTG is related to the Font Synthesis task, where the desired style must be represented and used to render characters consistently~\cite{azadi2018multi, cha2020few, park2021few, xie2021dg, lee2022arbitrary}. However, Font Synthesis approaches just need to generate single characters, thus, are more closely related to HTG for ideogrammatic languages~\cite{chang2018generating, gao2019artistic, jiang2019scfont,yuan2022se}. In general, both for ideogrammatic and non-ideogrammatic languages, handwriting can be treated either as a trajectory capturing the shape of the strokes making up the characters or as a static image capturing their overall appearance. Depending on this conception, online or offline approaches to HTG can be applied.

\tit{Online HTG} 
Approaches to online HTG exploit sequential models, such as LSTMs~\cite{graves2013generating}, Conditional Variational RNNs~\cite{aksan2018deepwriting}, or Stochastic Temporal CNNs~\cite{aksan2018stcn}, to predict the pen position point-by-point based on its current position and the input text to be rendered. The first approach following this strategy was proposed in~\cite{graves2013generating} and did not have control over the style. This limitation was then addressed by following works by decoupling and then recombining content and writer's style~\cite{aksan2018deepwriting, aksan2018stcn, kotani2020generating}. Further improvements to online HTG approaches can be obtained by training the sequential model alongside a discriminator~\cite{ji2019generative}, which is a philosophy similar to SotA GAN-based offline HTG approaches. The main drawbacks of approaches following the online HTG strategy are that they struggle to learn long-range dependencies and that they require training data consisting of digital pen recordings, which are difficult to collect or even impossible to obtain for application scenarios such as historical manuscripts. In the sight of these limitations, in this work, we follow the offline HTG paradigm.

\begin{figure*}[t]
    \centering
    \includegraphics[width=0.9\textwidth]{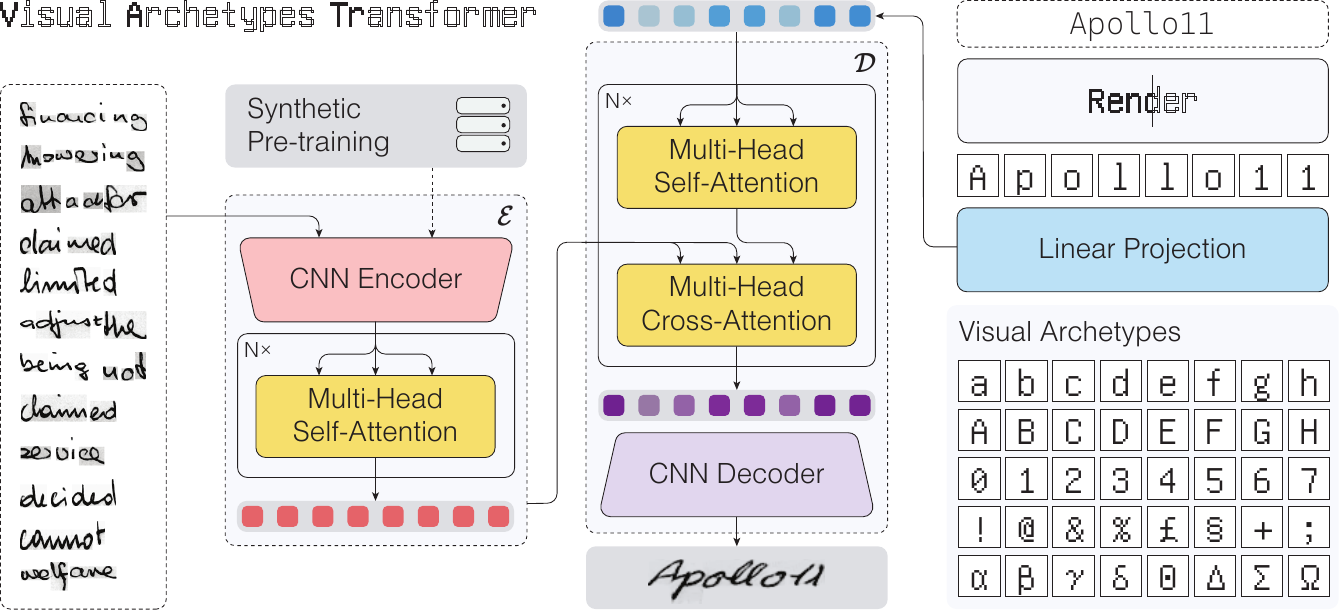}
    \caption{Overview of our Visual Archetypes-based Transformer for HTG (VATr). Few-shot learning is provided by a few images of the desired calligraphic style, encoded via a convolutional backbone pre-trained on a large synthetic dataset; the output vector is passed through a Transformer encoder for creating a latent space with robust style vectors (the  \emph{Style Encoder} $\mathcal{E}$ on the left). The text to generate is rendered as a sequence of GNU Unifont binary images, representing the visual archetypes of the characters. These are the queries of a Transformer decoder to perform cross-attention with the style vectors. The resulting content-style representation is then fed to a convolutional decoder that outputs the styled handwritten text image. These last two components are our \emph{Content-Guided Encoder} $\mathcal{D}$.\vspace{-.3cm}}
    \label{fig:overview}
\end{figure*}

\tit{Offline HTG} 
Traditional offline HTG solutions~\cite{wang2005combining, lin2007style, thomas2009synthetic, haines2016my} resort to heavy human intervention for glyphs segmentation and then apply handcrafted geometric static-based feature extraction before combining those glyphs with appropriate ligatures and rendering them with texture and background blending. Other than the costly human intervention that these approaches entail, their main limitation is that they can only render the glyphs and ligatures observed for each style. More recent deep learning-based approaches, instead, are able to infer styled glyphs even if not directly observed in the style examples.
Learning-based solutions rely on GANs, either unconditioned (for non-styled HTG) or conditioned on a varying number of handwriting style samples (for styled HTG). In this latter case, style samples can be entire paragraphs or lines~\cite{davis2020text}, a few words~\cite{kang2020ganwriting, bhunia2021handwriting}, or a single word~\cite{gan2021higan, gan2022higan+, luo2022slogan}. Collecting a few handwriting samples from a writer is not much more costly than one and generally results in better performance~\cite{krishnan2021textstylebrush}. The first of these approaches was proposed in~\cite{alonso2019adversarial} and was able to generate fixed-sized images conditioned on the content embedding but with no control over the calligraphic style. Note that, different from natural image generation, generating handwritten text images should entail producing variable-sized images. Thus, the approach presented in~\cite{fogel2020scrabblegan} aims at overcoming this limitation by concatenating character images, still not being able to imitate handwriting style. 

Approaches tackling styled HTG condition the generation not only on the text content but also on a vector representation of the style~\cite{davis2020text, kang2020ganwriting, mattick2021smartpatch, gan2021higan, gan2022higan+}. In such approaches, the style and the content representations are obtained separately and then combined in a later stage for generation. This prevents those approaches from effectively capturing local writing style and patterns. This limitation is addressed by the Transformer-based approach proposed in~\cite{bhunia2021handwriting}, which is able to better capture content-style entanglement by exploiting the cross-attention mechanism between the style vector representation and the content text representation. In this work, we follow the Transformer-based paradigm for its superior capability of rendering local style patterns.

\tit{Content Representation} 
The content tokens used in the approaches mentioned above~\cite{alonso2019adversarial, fogel2020scrabblegan, davis2020text, kang2020ganwriting, mattick2021smartpatch, gan2021higan, gan2022higan+, bhunia2021handwriting} are usually independent one-hot vectors, each representing a character in a finite and generally small charset. This strategy is thus limiting due to the relatively small charset that can be handled with reasonable computational cost and is inefficient for hindering the possibility of leveraging similarities between characters. Our approach, instead, is to exploit character images as text content inputs. Note that the approaches proposed in~\cite{krishnan2021textstylebrush, luo2022slogan} feed the textual input as a whole image containing the desired text written in a typeface font. These images are then rendered in the desired style in a style-transfer fashion, also exploiting the geometry encoded in the typeface-written image for letter spacing and curvature. Different from these approaches, we input text tokens as sequences of character images rendered in the richer GNU Unifont, which is as modular as the approaches employing one-hot encodings and allows exploiting geometric similarities between characters for generation.

\section{Proposed Approach}

The few-shot offline HTG problem that we tackle in this work can be formulated as follows. 
Consider a writer $\mathrm{w}{{\in}}\mathrm{W}$, for which we have $P$ samples of handwritten word images at disposal, $\mathbf{X}_{\mathrm{w}}{=}\{\mathbf{x}_{\mathrm{w},i}\}_{i=0}^P$ (in this work, following~\cite{kang2020ganwriting,bhunia2021handwriting}, we set $P{=}15$). Moreover, consider an arbitrarily long set of $Q$ text words $\mathbf{C}{=}\{\mathbf{c}_{i}\}_{i=0}^Q$, each containing an arbitrary number $n_{i}$ of characters. Our goal is to generate images $\mathbf{Y}_{\mathrm{w}}^{\mathbf{C}}$ of words with the content of the strings in $\mathbf{C}$ and the style of the writer $\mathrm{w}$ (see Figure~\ref{fig:first_page}, bottom).

\tit{Model Overview} 
We devise a Transformer encoder-decoder architecture, in combination with a pre-trained convolutional feature extractor for handling the style samples $\mathbf{X}_{\mathrm{w}}$, and rendered characters images for handling the content strings $\mathbf{C}$. First, a pre-trained convolutional feature extractor handles the style samples $\mathbf{X}_{\mathrm{w}}$ and feeds the resulting vectors to a Transformer encoder that enriches them with the long-range dependencies captured by the self-attention mechanism and outputs a sequence of style vectors $\mathbf{S}_{\mathrm{w}}$. The Transformer decoder performs cross-attention between $\mathbf{S}_{\mathrm{w}}$ and the content strings $\mathbf{C}$ to be rendered, which are represented as a sequence of their visual archetypes. The cross-attention mechanism brings to an entangled content-style representation that better captures local style patterns in addition to global word appearance. Finally, the obtained representation is fed into a convolutional decoder that generates the content and style conditioned word images $\mathbf{Y}_{\mathrm{w}}^{\mathbf{C}}$. We refer to this part of our architecture as the \emph{Content-Guided Decoder} $\mathcal{D}$. 
A schematic overview of our VATr architecture is reported in Figure~\ref{fig:overview}.

\subsection{Style Encoder}
The Style Encoder $\mathcal{E}$, which transforms the few sample images $\mathbf{X}_{\mathrm{w}}$ into the style features $\mathbf{S}_{\mathrm{w}}$, is a pipeline of a convolutional encoder and a Transformer encoder.
This choice is motivated by the data efficiency of convolutional neural networks and their ability to extract representative features and the suitability of the multi-head self-attention mechanism to model long-range dependencies in the style images.
The selected convolutional encoder backbone is a ResNet18~\cite{he2016deep}, which is a popular choice for approaches dealing with text images~\cite{javidi2020deep, zhu2020point, bhunia2021handwriting, manna2022swis}.  A novel additional characteristic is a pre-training process to obtain robust features from the style sample images. For this, we exploit a specifically built large dataset of word images rendered in calligraphic fonts. The details on the pre-training dataset are given in \textsection ~\ref{sssec:font2}.
Once pre-trained, we use the backbone to extract $P$ feature maps $\mathbf{h}_{\mathrm{w},i}{{\in}}\mathbb{R}^{h{\times}w{\times}d}$ from the $P$ style images $\mathbf{x}_{\mathrm{w},i}{{\in}}\mathbf{X}_{\mathrm{w}}$. These feature maps are then flattened along the spatial dimension to obtain a $(h{\cdot}w)$-long sequence of $d$-dimensional vectors. Note that while $h$ and $w$ depend on the input images shape, the embedding size $d$ is fixed and set equal to \num{512} in this work. The elements of this sequence represent adjacent regions of the original images, corresponding to the receptive field of the convolutional backbone.
The flattened feature maps of each style image are further concatenated to obtain the sequence $\mathbf{H}_{\mathrm{w}}{{\in}}\mathbb{R}^{N{\times}d}$, where $N{=}h{\cdot}w{\cdot}P$, which is fed into the first layer of the multi-layer multi-headed self-attention encoder. This encoder comprises $L{=}3$ layers, each with $J{=}8$ attention heads and a multilayer perceptron. The output of the last layer $\mathbf{H}^L{=}\mathbf{S}_{\mathrm{w}} {{\in}} \mathbb{R}^{N{\times}d}$ is the sequence of style features for writer $\mathrm{w}$, which is fed to the Transformer decoder in $\mathcal{D}$.

\subsubsection{Synthetic Pre-training}\label{sssec:font2}
Large-scale pre-training is an effective strategy employed in a number of learning tasks.
For HTG, the pre-training data should be abundant and should capture the shape variability of the glyphs and the texture characteristics of the ink and background.
In the sight of these considerations, to build the dataset used for pre-training the convolutional backbone, we render \numwords random words from the English vocabulary, each in \numfonts freely online available calligraphic fonts and on backgrounds randomly selected from a pool of paper-like images, thus obtaining more than 100M samples. 
To achieve better realism, we apply random transformations such as rotation and elastic deformation via the Thin Plate Spline transformation~\cite{duchon1977splines} to introduce shape variability, gaussian blur to avoid sharp borders and simulate handwriting strokes, and grayscale dilation and color jitter to simulate different ink types\footnote{Available at \url{https://github.com/aimagelab/VATr}}. 
We use the so obtained dataset to train the backbone to recognize the style of the word images by minimizing a Cross-Entropy Loss. Note that by exposing the network to such variability, we force it to extract features that are representative of the calligraphic style rather than the overall image appearance (which is influenced by the textual content, the background, and the ink type or writing tool).

\subsection{Content-Guided Decoder}
The first block of the \emph{Content-Guided Decoder} $\mathcal{D}$ is a multi-layer multi-head decoder with $L{=}3$ layers and $J{=}8$ heads as the encoder. 
The decoder performs self-attention between the content vectors representing the elements in $\mathbf{C}$, followed by cross-attention between the sequence of content vectors (treated as queries) and the style vectors $\mathbf{S}_{\mathrm{w}}$ (used as keys and values). In this way, the model can learn content-style entanglement since each query is forced to attend at the style vectors that are useful to render its final shape other than the general appearance. 

Unlike existing approaches that represent the content queries as embeddings of independent one-hot-encoded characters, in this work, we propose to exploit a representation that captures similarities between characters. In particular, we obtain the content queries as follows. 
Each content string $\mathbf{c}_{i} {{\in}} \mathbf{C}$ is made of a variable number of characters $k_i$, \ie~$\mathbf{c}_{i}{=}\{\mathbf{q}_{j}\}_{j=0}^{k_i}$. First, we render the characters in the GNU Unifont font, which, different from all other typeface fonts, contains all the Unicode characters. The rendering results in $16{\times}16$ binary images, which are then flattened and linearly projected to $d$-dimensional query embeddings, a strategy that is related to the direct use of image patches as input to Vision Transformer-like architectures~\cite{dosovitskiy2020image}. 

\begin{figure}
    \centering
    \includegraphics[width=\columnwidth]{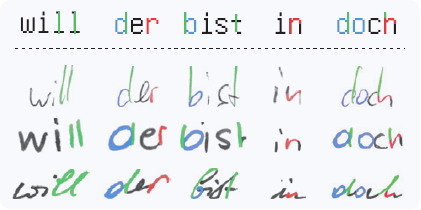}
    \caption{Comparison between the Unifont characters (top) and the same characters in different calligraphic styles (bottom). The geometric similarities between the characters are captured by the visual archetypes and thus can be exploited in generation.\vspace{-0.3cm}}
    \label{fig:embeddings}
\end{figure}

In Figure~\ref{fig:embeddings}, we show some exemplar visual archetypes (GNU Unifont characters) and the corresponding handwritten characters in different styles. It can be observed that the geometric similarities among the archetypes are reflected in styled characters. These similarities can be exploited for generating long-tail characters, \ie~characters that are rarely seen during training. In fact, by being fed with independent tokens as content queries, the network is forced to simply memorize content-shape relations. Not being exposed to a sufficient number of such pairs does not allow the model to learn such relations and results in unsatisfactory generation capabilities of long-tail characters. Conversely, with our image-based input, the network can learn to exploit geometric attributes and similarities between highly-represented and long-tail characters for rendering those latter, and thus, can generate them more faithfully. 
It is also worth noting that this character representation makes our model more scalable than one-hot encoding-based solutions. In fact, the query embedding layer we use has $256{\times}d$ parameters and allows us to handle a charset containing up to $2^{(16{\cdot}16)}$ characters. For handling the same amount of characters represented as one-hot vectors, the query embedding layer would have $2^{(16\cdot16)}{\times}d$ parameters.

The output of the last Transformer decoder layer for the content string $\mathbf{c}_{i}$ is a tensor $\mathbf{F}_{c_{i}}{\in}\mathbb{R}^{k_i{\times}d}$. We add normal gaussian noise to $\mathbf{F}_{c_{i}}$, to enhance variability in the generated images, and project it into a $(k_i{\times}\num{8192})$ matrix, which we then reshaped into a $512{\times}4{\times}4k_i$ tensor. This tensor is fed to a convolutional decoder consisting of four residual blocks and a $\mathrm{tanh}$ activation function that outputs the styled word images $\mathbf{Y}_{\mathrm{w}}^{\mathbf{C}}$.

\subsection{Model Training}
Formally, our complete VATr model is given by $\mathcal{G}_{\theta}=\mathcal{E}{\circ}\mathcal{D}:(\mathbf{X}_{\mathrm{w}},\textbf{C}){\rightarrow}\mathbf{Y}_{\mathrm{w}}^{\mathbf{C}}$. We train it alongside other modules used to calculate the overall loss for $\mathcal{G}_{\theta}$.

The first of those modules is a convolutional discriminator $\mathcal{D}_{\eta}$, which is trained to distinguish real images from images generated by $\mathcal{G}_{\theta}$, thus forcing the generator to produce realistic images. To optimize $\mathcal{G}_{\theta}$ and $\mathcal{D}_{\eta}$ we follow the adversarial paradigm with the hinge adversarial loss~\cite{lim2017geometric}
\begin{equation*}
\begin{split}
    L_{adv}=& \mathbb{E} \left[\text{max}(1 - \mathcal{D}_{\eta}(\mathbf{X}_{\mathrm{w}}), 0)\right] + \\
    & \mathbb{E} \left[\text{max}(1 + \mathcal{D}_{\eta}(\mathcal{G}_{\theta}(\mathbf{X}_{\mathrm{w}}, \textbf{C})), 0)\right].
\end{split}
\end{equation*}

Additionally, we exploit an HTR model~\cite{shi2016end}, $\mathcal{R}_{\phi}$, which is in charge of recognizing the text in the generated images, thus forcing the generator to reproduce the desired textual content other than rendering the style. The HTR model is trained with the real images $\mathbf{X}_{\mathrm{w}}$ and their ground truth transcription, while its loss value calculated on the generated images $\mathbf{Y}_{\mathrm{w}}^{\mathbf{C}}$ is propagated through the generator $\mathcal{G}_{\theta}$. The loss of the HTR model is obtained as
\begin{equation*}
    L_{HTR}=\mathbb{E}_{\mathbf{x}}\left[ - \sum \text{log}(p(t_{\mathbf{x}} | \mathcal{R}_{\phi}(\mathbf{x})))\right],
\end{equation*}
where $\mathbf{x}$ can be either a real or a generated image, and $t_{\mathbf{x}}$ is its transcription coming from the ground truth label in case $\mathbf{x}{\in}\mathbf{X}_{\mathrm{w}}$ or from $\mathbf{C}$ in case $\mathbf{x}{\in}\mathbf{Y}_{\mathrm{w}}^{\mathbf{C}}$.

Moreover, we employ a convolutional classifier $\mathcal{C}_{\psi}$ in charge of classifying the real and generated images based on their calligraphic style (\ie~the style of writer $\mathrm{w}$), thus forcing the generator $\mathcal{G}_{\theta}$ to render the correct style. As done for the $\mathcal{R}_{\phi}$ module, also this classifier is trained with the real images, and its loss value on the generated images is used to guide the generator. Formally, the loss for this module is
\begin{equation*}
    L_{class}=\mathbb{E}_{\mathbf{x}}\left[ - \sum \text{log}(p(\mathrm{w} | \mathcal{C}_{\psi}(\mathbf{x})))\right].
\end{equation*}
Also in this case, $\mathbf{x}{\in}\mathbf{X}_{\mathrm{w}}$ or $\mathbf{x}{\in}\mathbf{Y}_{\mathrm{w}}^{\mathbf{C}}$.

To further enforce the generation of images in the desired style, we use an additional regularization loss, namely the cycle consistency loss given by:
\begin{equation*}
    L_{cycle}=\mathbb{E} \left[ \left\lVert \mathcal{E}(\mathbf{X}_{\mathrm{w}}) - \mathcal{E}(\mathbf{Y}_{\mathrm{w}}^{\mathbf{C}}) \right\rVert_1 \right].
\end{equation*}
The rationale is to force the generator to produce styled images for which the encoder $\mathcal{E}$ would extract the same style vectors. In other words, we want the style features of the input images to be preserved in the generated ones.

Overall, the complete objective function we use to train our model is given by combining the above loss terms equally weighed, \ie
\begin{equation*}
    L = L_{adv} + L_{HTR} +  L_{class} + L_{cycle}.
\end{equation*}
For an analysis of the role of each loss term on the performance, we refer to the Supplementary material.
\begin{figure}
    \centering
    \includegraphics[width=\columnwidth]{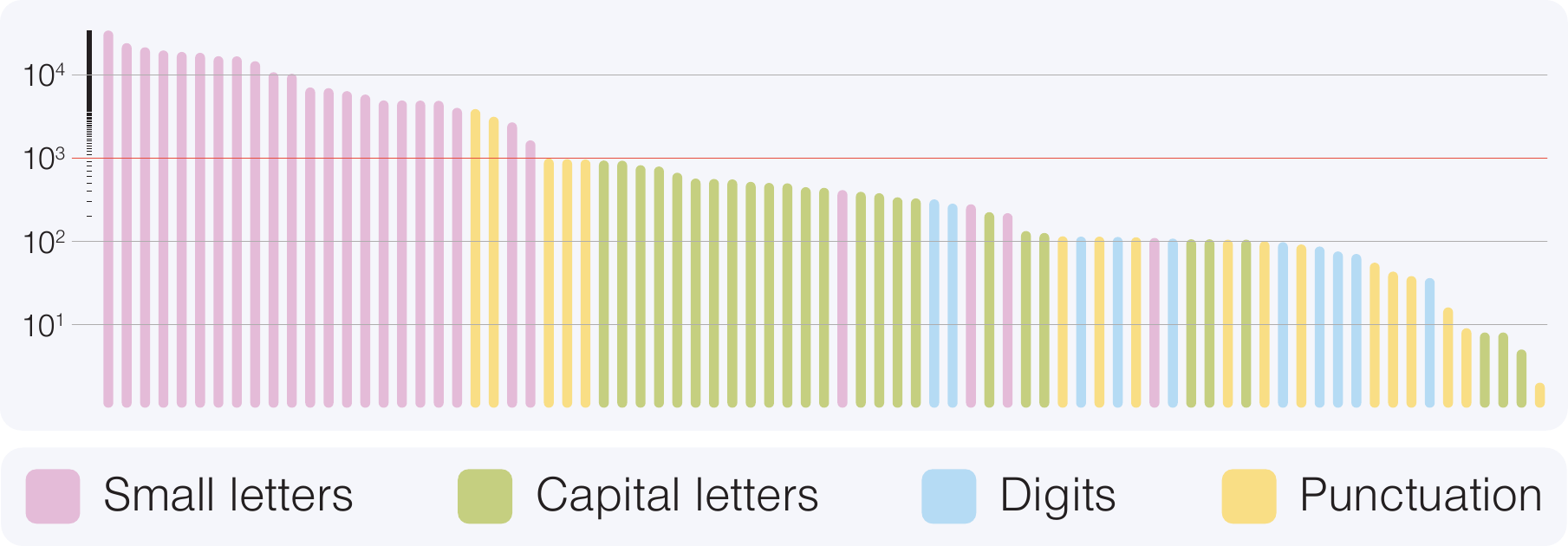} 
    \caption{Distribution and classification of the characters in the training set of the IAM dataset (in logarithmic scale). We set a threshold on the frequency with which the characters appear equal to \num{1000} to identify the long-tail ones (indicated as a red line).\vspace{-.2cm}}
    \label{fig:iam_class_tail}
\end{figure} 

\section{Experiments}
\label{sec:experiments}

In this section, in addition to analyzing the performance in the standard styled HTG scenarios, we aim to explore the capability of our approach and of SotA ones to generate characters that are long-tail-distributed in the dataset used for training. We believe that this is a relevant aspect to consider when evaluating the HTG performance, which has been so far neglected in the literature on this task. 
Additional analysis is reported in supplementary materials.

\tit{Implementation Details}
All the experiments have been carried out on a single NVIDIA RTX 2080~Ti GPU. For pre-training the convolutional style encoder, we set the batch size to \num{32}. We use the Adam optimizer with an initial learning rate equal to $2 \times 10^-5$ and apply exponential scheduling with a decay factor equal to $10^{-1/90000}$. We stop the training with an early stopping strategy with patience \num{30}. Note that, due to the large amount of samples in the dataset, we are able to feed the convolutional backbone with almost always unseen samples before convergence. For this reason, we count an epoch every \num{1000} iterations. We employ the Adam optimizer also for training the complete HTG model on the real benchmark dataset considered in this work but fix the learning rate to $2\cdot10^{-4}$ and batch size equal to \num{8}. In this case, the training is stopped after \num{7}k epochs.

\tit{Evaluation Protocol}
For validating our proposal and comparing it as fairly as possible against SotA HTG approaches, we consider the widely-used IAM dataset~\cite{marti2002iam}, with images rescaled to have \num{64} pixels in height and proportional width. Moreover, we adopt the same evaluation procedure as in~\cite{kang2020ganwriting, bhunia2021handwriting}. In particular, the dataset contains \num{62857} English words from the Lancaster-Oslo/Bergen (LOB) corpus~\cite{johansson1978manual}, handwritten by multiple users. For this work, we consider the words written by \num{340} of those users for training and those written by the remaining \num{160} for testing. The words in the dataset are composed starting from an alphabet of \num{79} characters distributed in the training set as shown in Figure~\ref{fig:iam_class_tail}. It can be noticed that these characters appear in the dataset in a long-tail distribution: small letters are the most represented (note that `e', `t', `a', and `o' are the most common letters, which reflects the frequency in the English vocabulary), while almost all the capital letters, digits, and punctuation are rare characters in the dataset. In our experiments, we consider long-tail characters those appearing less than \num{1000} times in the training set.
We compare the proposed VATr model against the following learning-based methods for HTG. When available, we use the official implementation and weights released by the authors and evaluate all the models in the same setups. In particular, we consider the non-styled HTG ScrabbleGAN~\cite{fogel2020scrabblegan}, the one-shot styled HTG model HiGAN~\cite{gan2021higan}, the approach proposed in~\cite{davis2020text} (which we refer to as TS-GAN), and the few-shot styled methods GANwriting~\cite{kang2020ganwriting} and HWT~\cite{bhunia2021handwriting}. These two latter approaches use the same number of style examples as we use. Notably, HWT is the most closely related to our proposal for following a Transformer encoder-decoder paradigm for HTG. 
For performance evaluation, we consider the Fréchet Inception Distance (FID)~\cite{heusel2017gans} and the Geometry Score (GS)~\cite{khrulkov2018geometry} for measuring the visual quality of the generated images. Moreover, to further evaluate the capability of the considered methods to generate rare characters faithfully, we calculate the Character Error Rate (CER) of an HTR network trained on the IAM dataset~\cite{shi2016end} when recognizing the text in the generated images. Note that, for all the scores, the lower the value, the better.

\begin{figure*}[t]
    \centering
    \includegraphics[width=\textwidth]{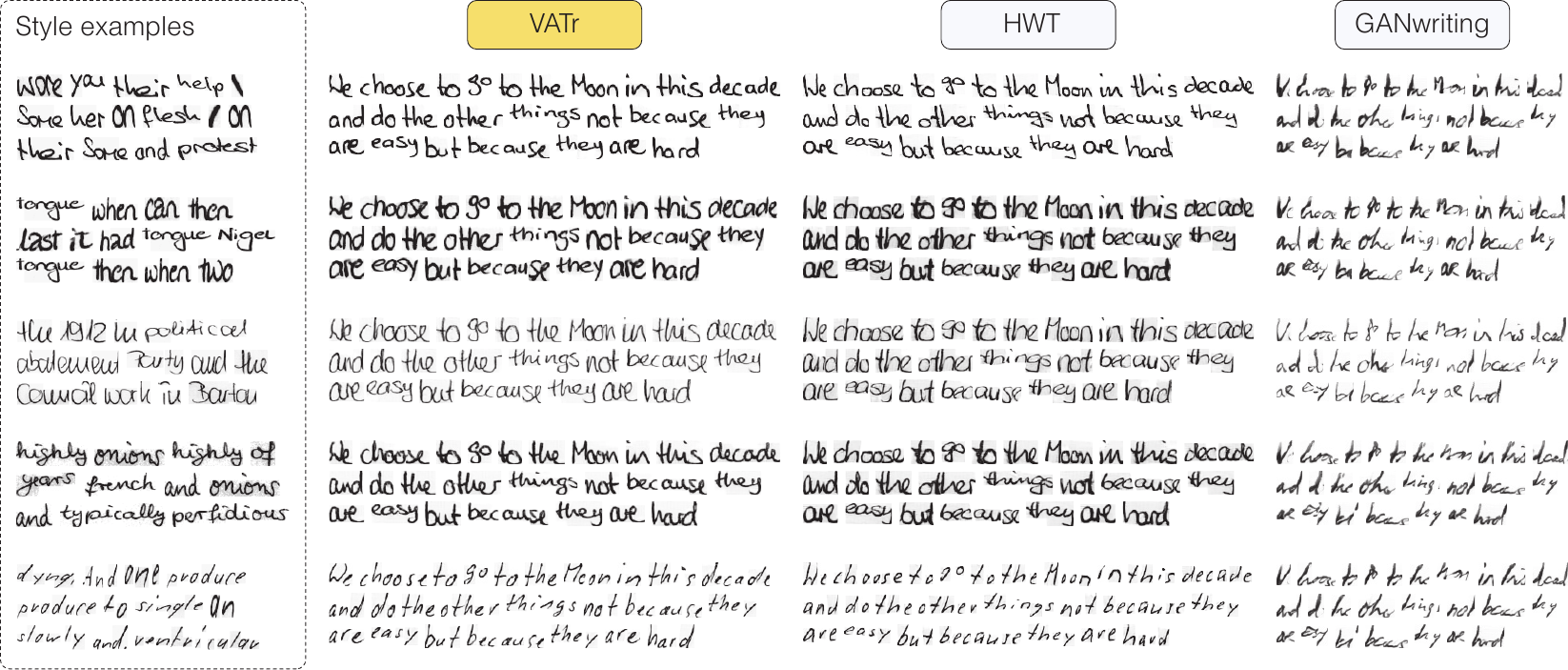}
    \caption{Qualitative comparison between our approach and the few-shot style HTG competitors in generating images with the desired textual content in the desired calligraphic style.\vspace{-0.2cm}}
    \label{fig:overall_qualitatives}
\end{figure*}

\subsection{Ablation Analysis}
First, we validate the benefits of using visual archetypes instead of one-hot vectors by replacing those latter as input to $\mathcal{D}$. The results of this analysis are reported in Table~\ref{tab:components_abl}, both when generating styled words in the whole IAM test set and when generating words containing long-tail characters. It can be observed that our approach is superior to the baseline, especially for the generation of long-tail words.
Moreover, we study the effect of the proposed synthetic pre-training strategy by comparing the performance obtained when training on real images only, and when pre-training to recognize the writers in the IAM dataset. Also these results are reported in Table~\ref{tab:components_abl} and show that the proposed synthetic pre-training brings more gain than training on real data, especially when used in combination with visual archetypes.

\begin{table}[]
    \footnotesize
    \centering
    \setlength{\tabcolsep}{.38em}
    \caption{Ablation analysis on the components of VATr.}
    \label{tab:components_abl}
    \resizebox{\linewidth}{!}{
    \begin{tabular}{c c c c}
    \toprule
    \rule{0pt}{-10pt}\textbf{Pre-training} & \textbf{Content input} & \textbf{FID (All)} & \textbf{FID (Long-Tail)}\\
    \midrule
    None         & One-hot vectors   & 18.48 & 24.93 \\ 
    Synthetic    & One-hot vectors   & 19.19 & 23.71 \\ 
    \midrule
    None         & Visual archetypes & 17.91 & 22.15 \\ 
    IAM          & Visual archetypes & 18.93 & 21.88 \\ 
    Synthetic    & Visual archetypes & 17.79 & 21.36 \\ 
    \bottomrule
    \end{tabular}}\vspace{-0.3cm}
\end{table}

\begin{table}[t]
\footnotesize
\centering
\setlength{\tabcolsep}{.32em}
\caption{Generated image quality evaluation on the IAM test set, regardless of the calligraphic style. Best results in bold.}
\label{tab:overall}
\resizebox{0.72\linewidth}{!}{
\begin{tabular}{l c cc}
\toprule
 && FID & GS\\
\midrule
ScrabbleGAN\cite{fogel2020scrabblegan}&& 20.72 & 2.56$\times$10$^{\text{-2}}$\\
HiGAN\cite{gan2021higan}&& 24.90 & 3.19$\times$10$^{\text{-2}}$\\  
TS-GAN\cite{davis2020text}&& 20.65 & 4.88$\times$10$^{\text{-2}}$\\
HWT\cite{bhunia2021handwriting}&& 19.40 & \textbf{1.01$\times$10$^{\text{-2}}$}\\
\textbf{VATr (Ours)} && \textbf{17.79} & 1.68$\times$10$^{\text{-2}}$\\
\bottomrule
\end{tabular}
}\vspace{-.2cm}
\end{table}

\subsection{Few-Shot Styled HTG}
\begin{figure}[t]
    \centering
    \includegraphics[width=\columnwidth]{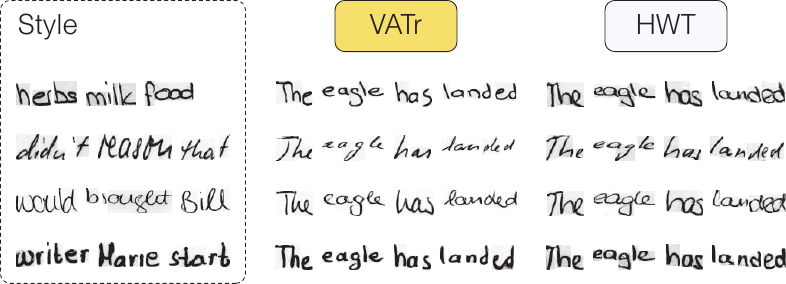}
    \caption{Exemplar generated images from style images with background artifacts.
    \vspace{-.4cm}}
    \label{fig:background_qualitatives}
\end{figure}

In this section, we evaluate the capability of the proposed approach to generate realistic handwritten text images, regardless of the calligraphic style. To this end, we calculate the FID and GS on the IAM test set. The results of this study are reported in Table~\ref{tab:overall}. It can be observed that our approach gives the best FID score and is second-best in terms of GS by a small margin, suggesting the realism of its generated images. 
As for the styled HTG evaluation, we follow the procedure proposed in~\cite{kang2020ganwriting}. In particular, we calculate the FID of the generated images in comparison to the real ones for each considered writer separately and then average the scores. We perform our analysis by distinguishing four increasingly challenging scenarios, namely: 1) the IV-S case, in which we generate in-vocabulary words in styles seen during training (\ie~both style and content have been seen during training); 2) the IV-U case, in which the words to generate are in-vocabulary, but the style has never been observed; 3) the OOV-S case, in which the textual content consists of out-of-vocabulary words, but the style has been encountered during training; 4) the OOV-U case, in which both the desired style and words are unseen. The results of this analysis are reported in Table~\ref{tab:iam_scenarios}. It can be observed that our approach outperforms the competitors in all four settings by a large margin. 
Some qualitative results are reported in Figure~\ref{fig:overall_qualitatives}, which refer to the generation of a text with different unseen styles.
It is also worth noting that, thanks to our large-scale synthetic pre-training strategy, VATr is able to focus on the shape attributes of the style to reproduce rather than on the background. This results in clearer generated images that reflect the handwriting of the reference ones rather than nuisances in the background (see Figure~\ref{fig:background_qualitatives}). Note that separating the handwriting from the background can negatively affect the FID score, which works on Inception v3 features, and even more the GS since it is calculated directly on the images. 
Nevertheless, this is an interesting capability for HTG models since it makes them suitable for generating styled text that can be easily superimposed to any desired background without artifacts.

\begin{table}[t]
\footnotesize
\centering
\setlength{\tabcolsep}{.32em}
\caption{Generated image quality evaluation by considering seen and unseen calligraphic style and in-vocabulary and out-of-vocabulary textual content. Best results in bold.}
\label{tab:iam_scenarios}
\resizebox{\linewidth}{!}{
\begin{tabular}{l c c c cc }
\toprule
 && IV-S & IV-U & OOV-S & OOV-U \\
\midrule
TS-GAN\cite{davis2020text} && 118.56 & 128.75 & 127.11 & 136.67\\
GANwriting\cite{kang2020ganwriting} && 120.07 & 124.30 & 125.87 & 130.68\\
HWT\cite{bhunia2021handwriting} && 106.97 & 108.84 & 109.45 & 114.10\\
\textbf{VATr (Ours)} && \textbf{88.20} & \textbf{91.11} & \textbf{98.57} & \textbf{102.22}\\
\bottomrule
\end{tabular}
}\vspace{.3cm}
\end{table}

\begin{figure}[]
    \centering
    \includegraphics[width=\columnwidth]{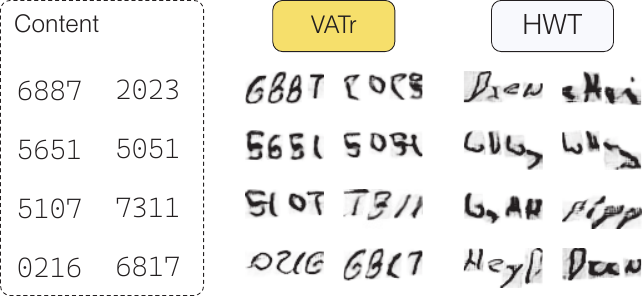}
    \caption{Comparison of the images of numbers generated by our approach and HWT.
    \vspace{-.3cm}}
    \label{fig:numbers_qualitatives}
\end{figure}

\subsection{Long-Tail Characters Generation}
In this section, we focus on the capability of our visual archetypes-based approach to faithfully render rare characters. Note that the split of the IAM dataset used in training contains a total of \num{66608} word images. Among those, only \num{13064} contain at least one long-tail character. In Table~\ref{tab:longtail}, we present the performance on the generation of test strings that contain those characters, with a further evaluation on words made up of just digits. In particular, we evaluate both style and content preservation by measuring the FID and the CER. 
As can be observed from the values of the FID, SotA approaches relying on one-hot vector encodings of the content struggle to generate realistic images, especially when these contain only rare characters, as in the case of numbers. Our approach, instead, can handle such words more easily by exploiting shape similarity between the visual archetypes of the characters to render. This is confirmed by the qualitative results of the generation of numbers reported in Figure~\ref{fig:numbers_qualitatives}, where we compare our approach against HTW to better highlight the benefit of using the visual archetypes over one-hot encodings. It can be observed that the images generated by HWT do not resemble digits, while those of VATr better preserve the content.

\begin{figure}[t]
    \centering
    \includegraphics[width=\columnwidth]{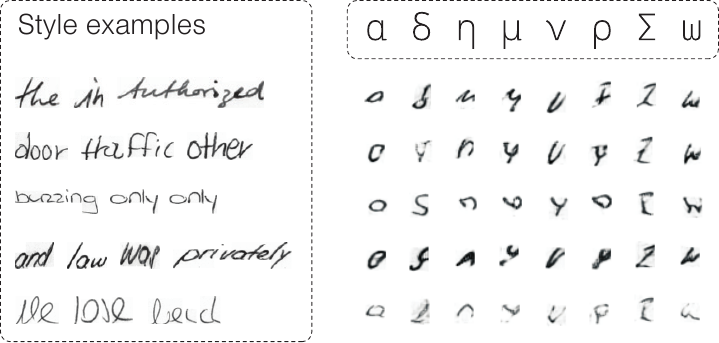}
    \caption{Generated images of some out-of-charset symbols (greek letters) in different styles.}
    \label{fig:out_of_charset}
\end{figure}

Finally, it is worth mentioning that our approach comes with the machinery to generate, to some extent, also out-of-charset characters, \ie~unseen symbols (\eg~from other alphabets) in different handwriting styles. In particular, when those unseen symbols share visual details with the characters encountered during training (\eg~as in the case of Greek letters `$\delta$' and `$\omega$' and Latin letters `s' and `w'), our model can resort to the geometric patterns learned for the latter. Some qualitative examples of out-of-charset generation are given in Figure~\ref{fig:out_of_charset}. Although the visual quality is inferior compared to that of the seen characters, our VATr strives to generate some out-of-charset symbols.

\begin{table}[t]
\footnotesize
\centering
\setlength{\tabcolsep}{.32em}
\caption{Generated image quality evaluation by considering words containing at least one among the long-tail characters in the IAM dataset, and just numbers. The CER value calculated on real images is reported for reference. Best results in bold.}
\label{tab:longtail}
\resizebox{0.85\linewidth}{!}{
\begin{tabular}{l c cc c cc}
\toprule
 && \multicolumn{2}{c}{All Long-Tail} && \multicolumn{2}{c}{Digits}\\
 \cmidrule{3-4} \cmidrule{6-7}
 && FID & CER && FID & CER \\
\midrule
Real Images && - & 6.21 && - & 45.80\\
\midrule
HiGAN\cite{gan2021higan}&& 26.08 & \textbf{8.63} && 129.61 & 101.53\\
HWT\cite{bhunia2021handwriting}&& 40.95 & 20.36 && 131.74 & 98.47\\
\textbf{VATr (Ours)} && \textbf{21.36} & 11.85 && \textbf{104.12} & \textbf{94.66}\\
\bottomrule
\end{tabular}
}\vspace{.3cm}
\end{table}

\section{Conclusion}
In this work, we have proposed VATr, a few-shot styled HTG approach that is able to reproduce unseen calligraphic styles and generate characters rarely encountered in the training set. These capabilities are achieved by exploiting supervised pre-training on a large synthetic dataset of calligraphic fonts and by representing the textual content as a sequence of visual archetypes, \ie~binary images of Unifont-rendered characters. Experimental results demonstrate that by pre-training, we are able to extract more representative style features that disregard the background and the ink texture. Moreover, by using the visual archetypes, we are able to exploit shape similarities among characters, which eases the generation of the rare ones.

\section*{Acknowledgement}
This work was supported by the ``AI for Digital Humanities'' project (Pratica Sime n.2018.0390), funded by ``Fondazione di Modena'' and the PNRR project Italian Strengthening of Esfri RI Resilience (ITSERR) funded by the European Union – NextGenerationEU (CUP: B53C22001770006).

{\small
\balance
\bibliographystyle{ieee_fullname}
\bibliography{main}
}

\end{document}


\title{Handwritten Text Generation from Visual Archetypes\\Supplementary Material}

\author{Vittorio Pippi, Silvia Cascianelli, Rita Cucchiara\\
University of Modena and Reggio Emilia\\
Via Pietro Vivarelli, 10, Modena (Italy)\\
{\tt\small \{name.surname\}@unimore.it}
}
\maketitle

In this document, we present additional experimental analysis of our proposed approach for Few-Shot HTG, VATr. In particular, we show 
an ablation analysis on the role of each of the used loss terms in \textsection\ref{sec:losses}; 
additional details about the synthetic pre-training dataset in \textsection~\ref{sec:dataset}; 
additional qualitative results in \textsection~\ref{sec:qualitatives}; 
a comparison of the charsets generated in different styles in \textsection~\ref{sec:charsets}; 
further analysis of the long-tail characters generation scenario in \textsection~\ref{sec:long-tail}; 
examples of the generation of some out-of-charset characters from other alphabets in \textsection\ref{sec:outofcharset}; 
and some HTR results obtained by training both on real and VATr-generated images in \textsection\ref{sec:outofcharset}.

\section{Role of the Loss Function Terms}\label{sec:losses}
We analyze the effect of each loss term, both quantitatively and qualitatively. The results of this study are reported in Table~\ref{tab:losses_abl} and show that our model needs the included regularization terms to be trained properly. Moreover, when none of the two terms enforcing handwriting style faithfulness (\ie~the writer classification loss $L_{class}$ and the cycle loss $L_{cycle}$) is used, VATr generates images of readable words but does not render the style of the reference images. Conversely, when the text recognition loss $L_{HTR}$ is not employed, it renders "scrabbles" whose overall appearance resembles that of the reference style images.

\section{Synthetic Pre-training Dataset}\label{sec:dataset}
\begin{figure}
    \centering
    \includegraphics[width=\columnwidth]{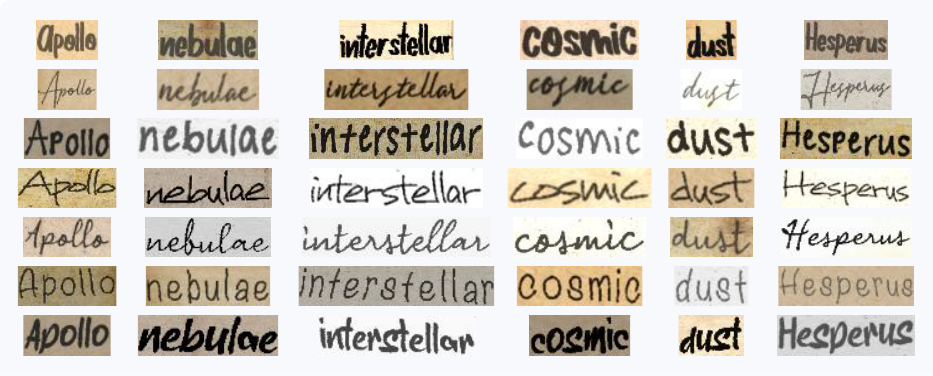} 
    \caption{Exemplar images from the synthetic dataset used for pre-training the style encoder.}
    \label{fig:font2}
\end{figure}

The synthetic dataset used to pre-train the convolutional backbone of the encoder is obtained by combining calligraphic fonts and words to generate word images. The process involves selecting \num{10400} fonts from dedicated websites, ensuring that fonts with small caps and decorative elements such as hearts or stars are discarded. Next, \num{10400} words of varying lengths are randomly chosen from the English vocabulary. All possible combinations between the selected fonts and words are then rendered, resulting in a total of $\num{10400} \times \num{10400} = \num{108160000}$ word images. Some exemplar images are reported in Figure~\ref{fig:font2}.

\section{Additional Styled HTG Qualitative Results}\label{sec:qualitatives}
\begin{figure*}
    \textbf{IV-S scenario} \\ \\
    \vspace{0.5cm}
    \includegraphics[width=\textwidth]{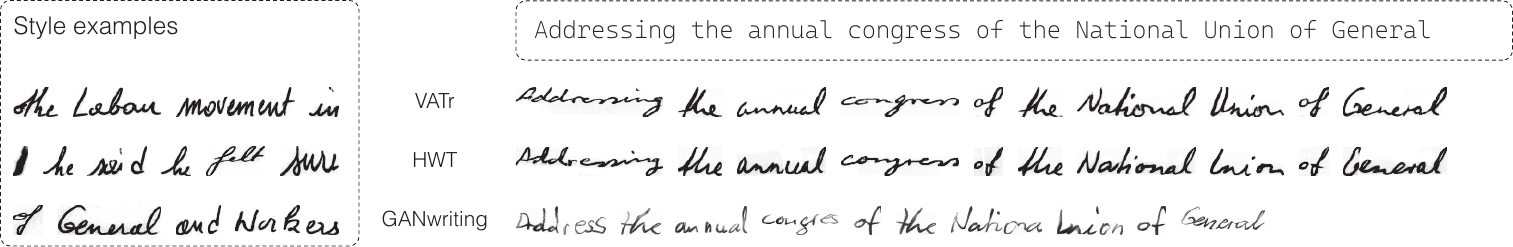} \\
    \vspace{0.5cm}
    \includegraphics[width=\textwidth]{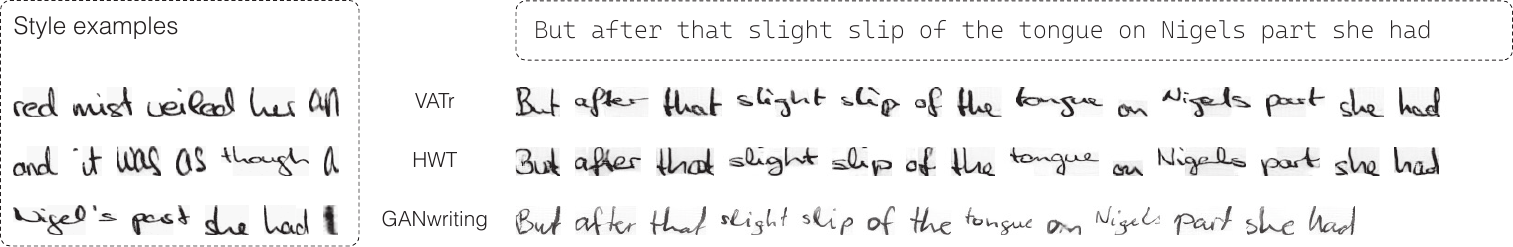} \\
    \vspace{0.5cm}
    \includegraphics[width=\textwidth]{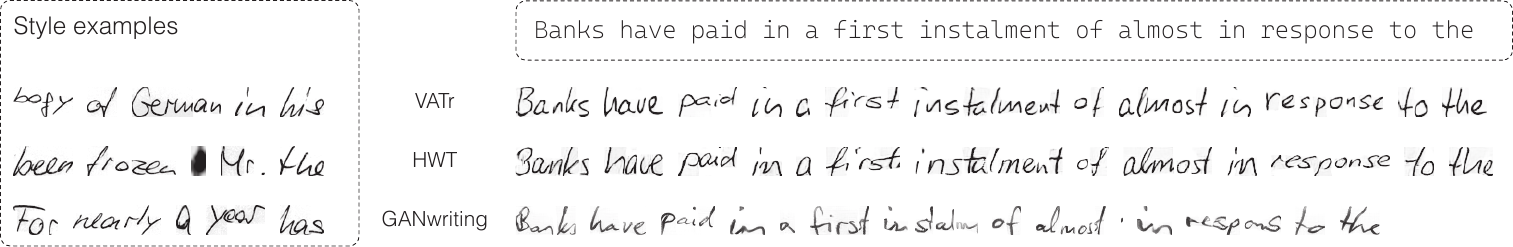} \\
    \vspace{0.5cm}
    \includegraphics[width=\textwidth]{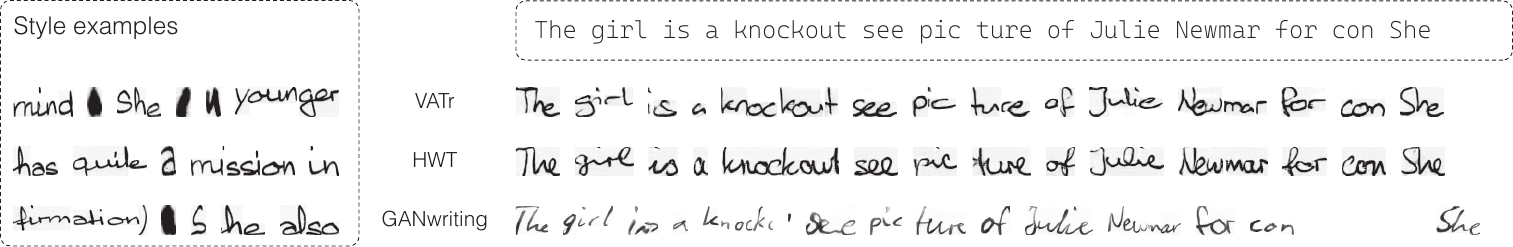} \\
    \vspace{0.5cm}
    \includegraphics[width=\textwidth]{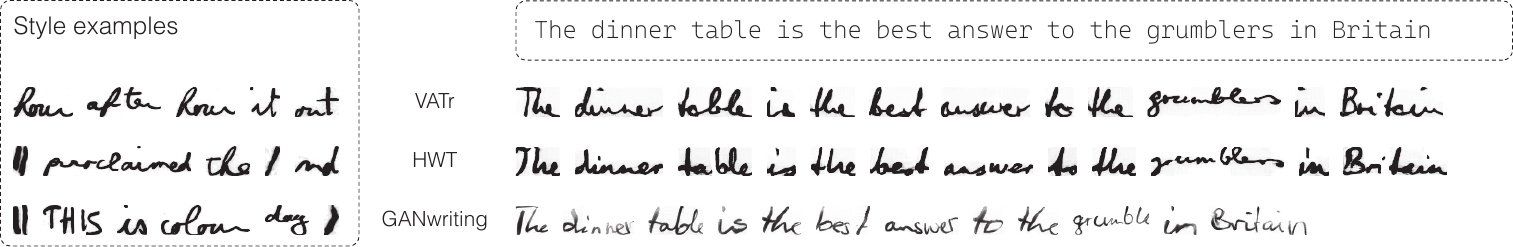} \\
    \vspace{0.5cm}
    \includegraphics[width=\textwidth]{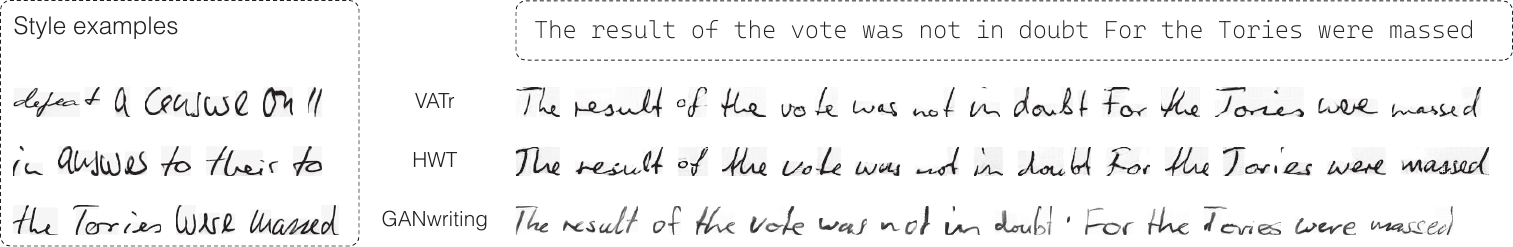} 
    \caption{Exemplar qualitative results of styled text generation for the IV-S setting on the IAM dataset.}
    \label{fig:iam_iv_s}
\end{figure*}

\begin{figure*}
    \textbf{IV-U scenario} \\ \\
    \vspace{0.5cm}
    \includegraphics[width=\textwidth]{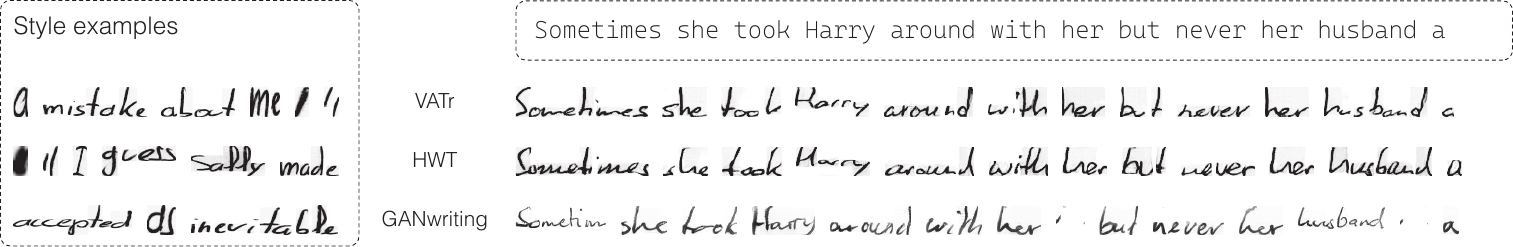} \\
    \vspace{0.5cm}
    \includegraphics[width=\textwidth]{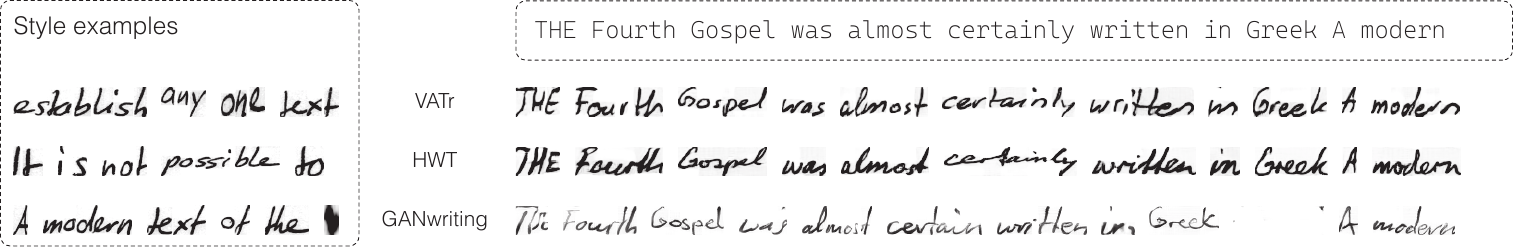} \\
    \vspace{0.5cm}
    \includegraphics[width=\textwidth]{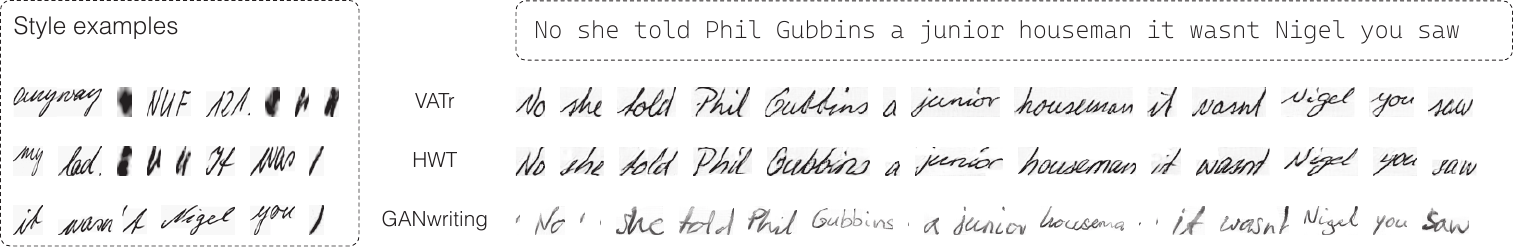} \\
    \vspace{0.5cm}
    \includegraphics[width=\textwidth]{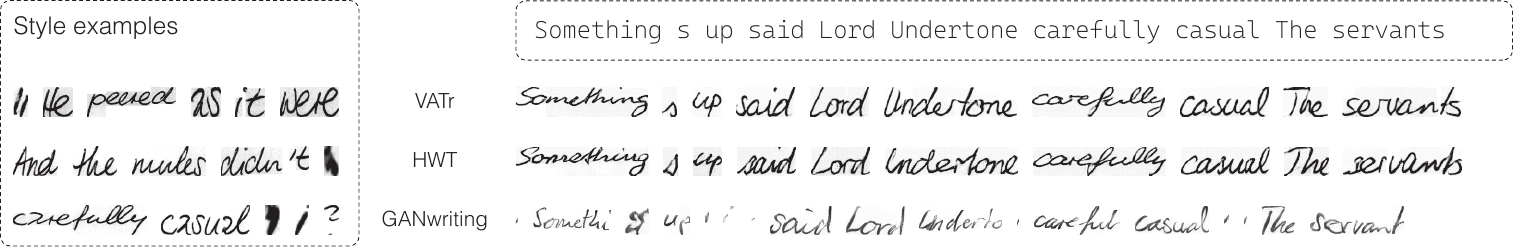} \\
    \vspace{0.5cm}
    \includegraphics[width=\textwidth]{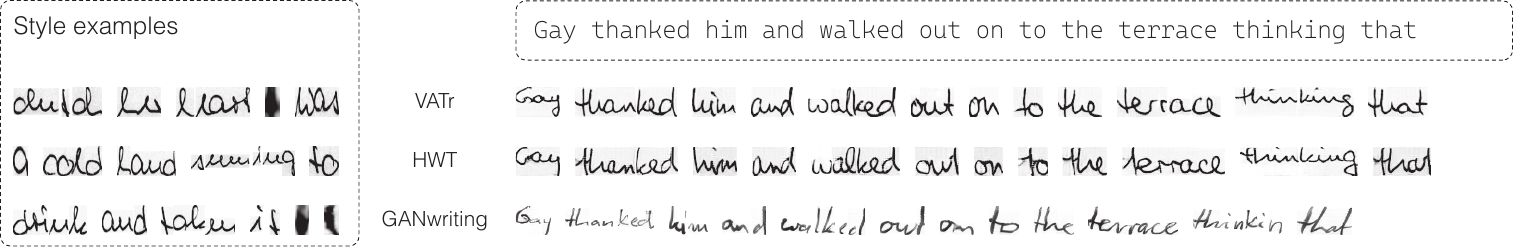} \\
    \vspace{0.5cm}
    \includegraphics[width=\textwidth]{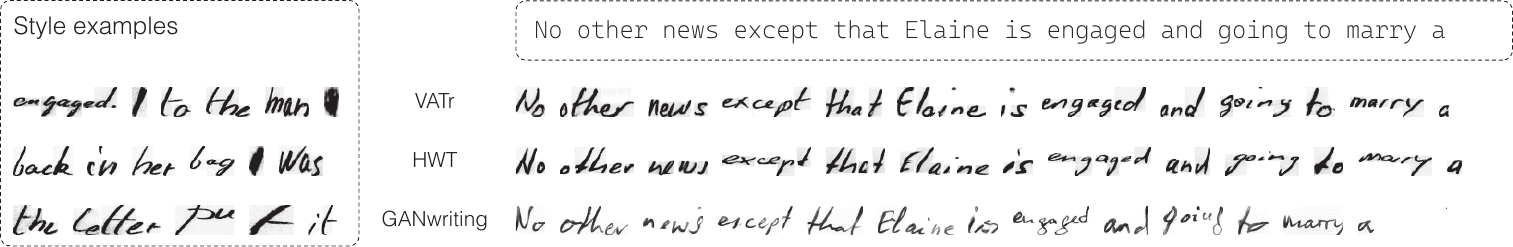} 
    \caption{Exemplar qualitative results of styled text generation for the IV-U setting on the IAM dataset.}
    \label{fig:iam_iv_u}
\end{figure*}

\begin{figure*}
    \textbf{OOV-S scenario} \\ \\
    \vspace{0.5cm}
    \includegraphics[width=\textwidth]{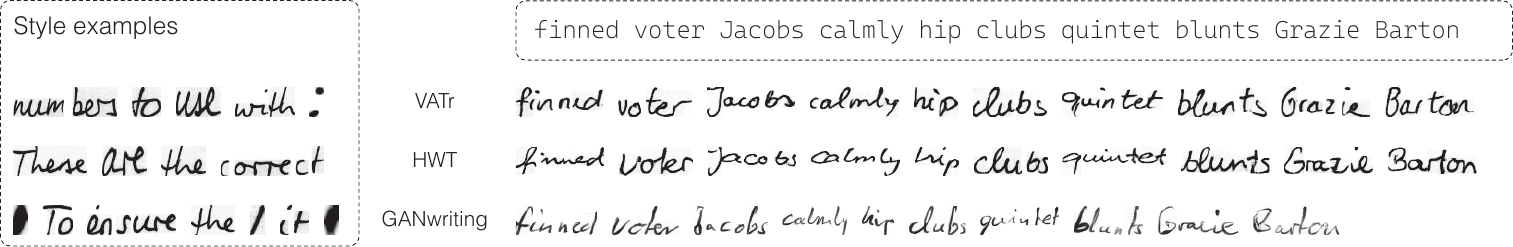} \\
    \vspace{0.5cm}
    \includegraphics[width=\textwidth]{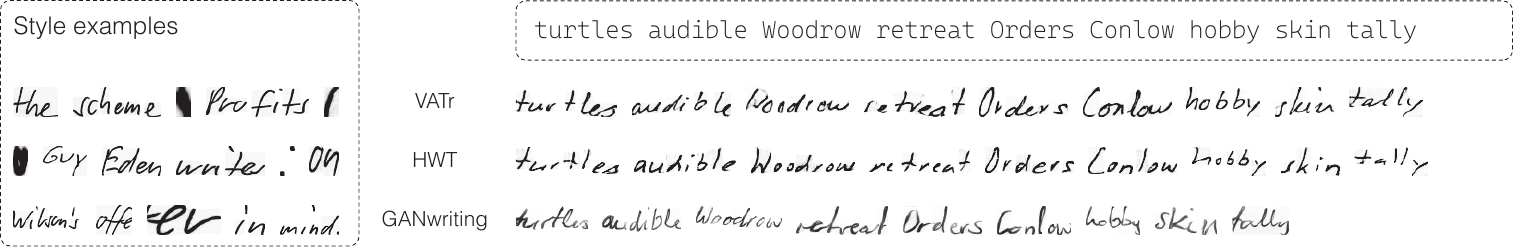} \\
    \vspace{0.5cm}
    \includegraphics[width=\textwidth]{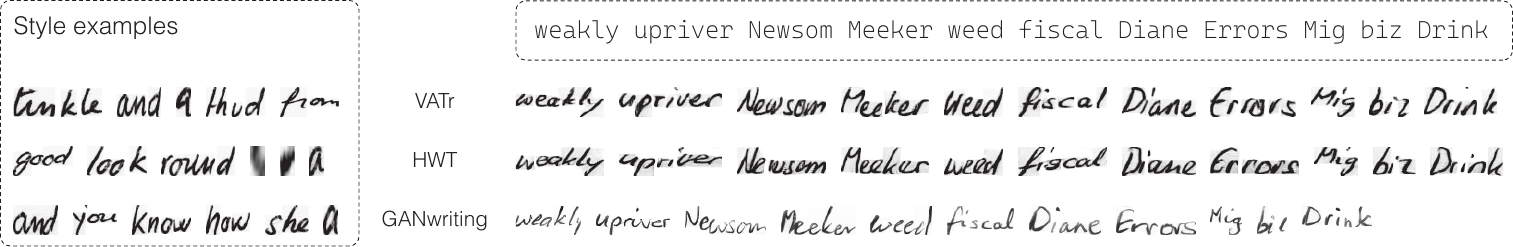} \\
    \vspace{0.5cm}
    \includegraphics[width=\textwidth]{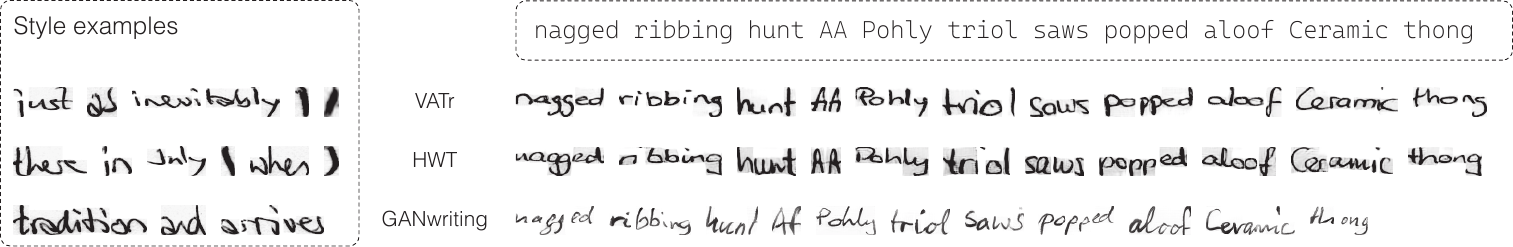} \\
    \vspace{0.5cm}
    \includegraphics[width=\textwidth]{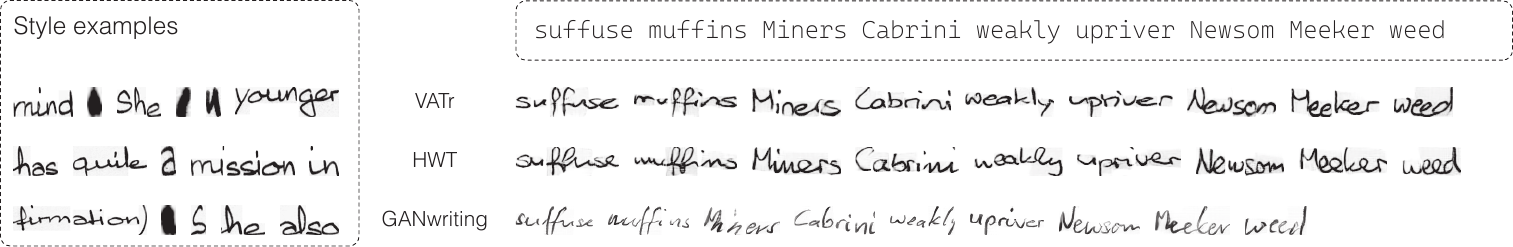} \\
    \vspace{0.5cm}
    \includegraphics[width=\textwidth]{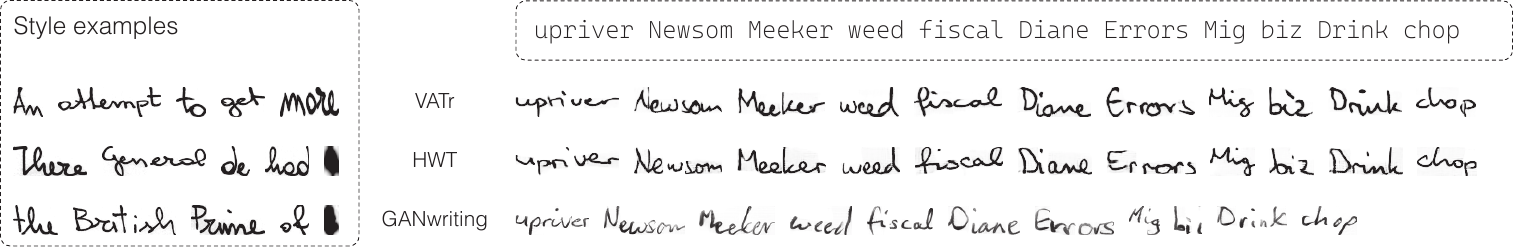} 
    \caption{Exemplar qualitative results of styled text generation for the OOV-S setting on the IAM dataset.}
    \label{fig:iam_oov_s}
\end{figure*}

\begin{figure*}
    \textbf{OOV-U scenario} \\ \\
    \vspace{0.5cm}
    \includegraphics[width=\textwidth]{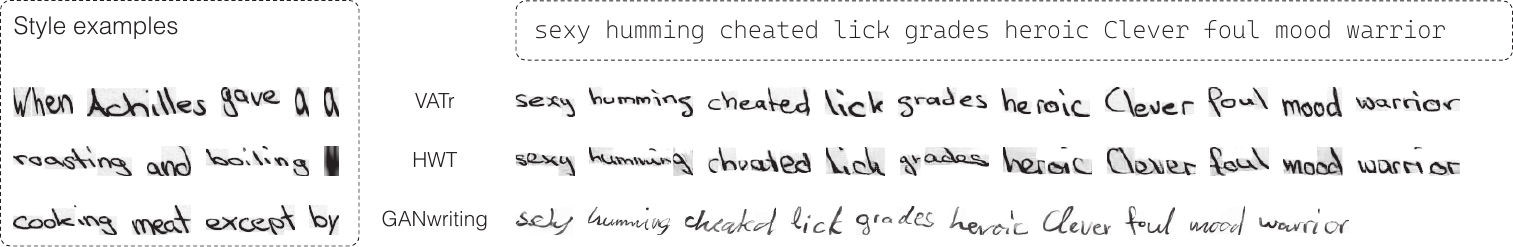} \\
    \vspace{0.5cm}
    \includegraphics[width=\textwidth]{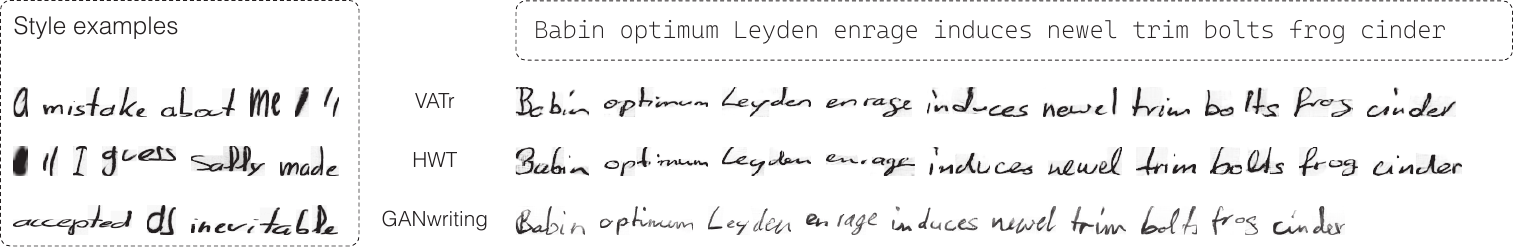} \\
    \vspace{0.5cm}
    \includegraphics[width=\textwidth]{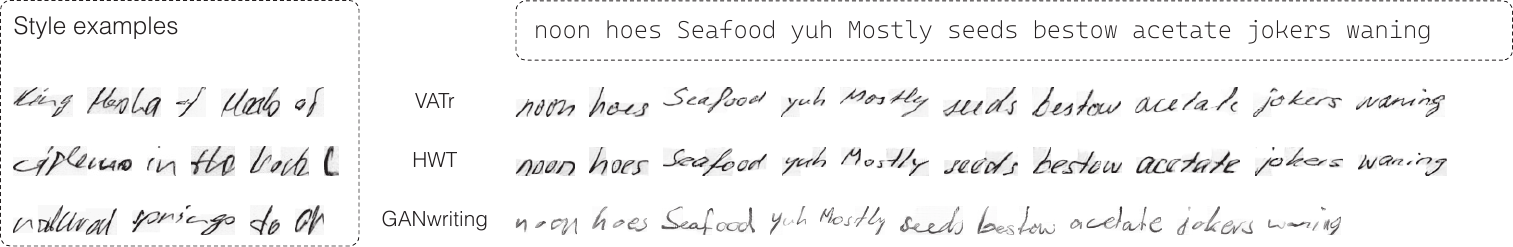} \\
    \vspace{0.5cm}
    \includegraphics[width=\textwidth]{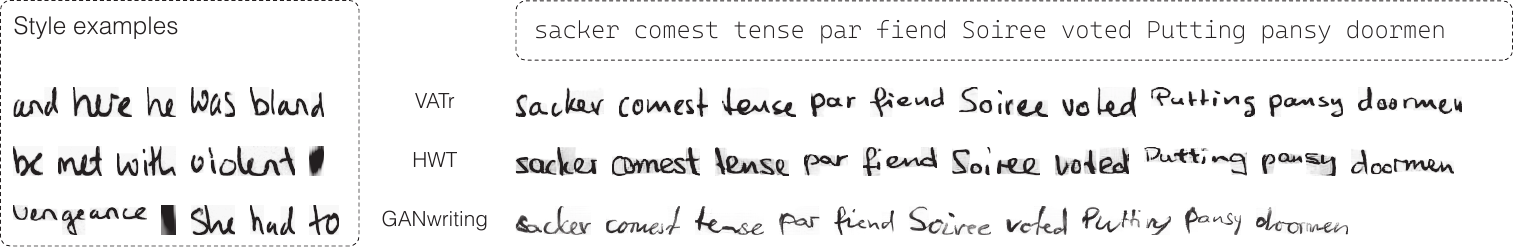} \\
    \vspace{0.5cm}
    \includegraphics[width=\textwidth]{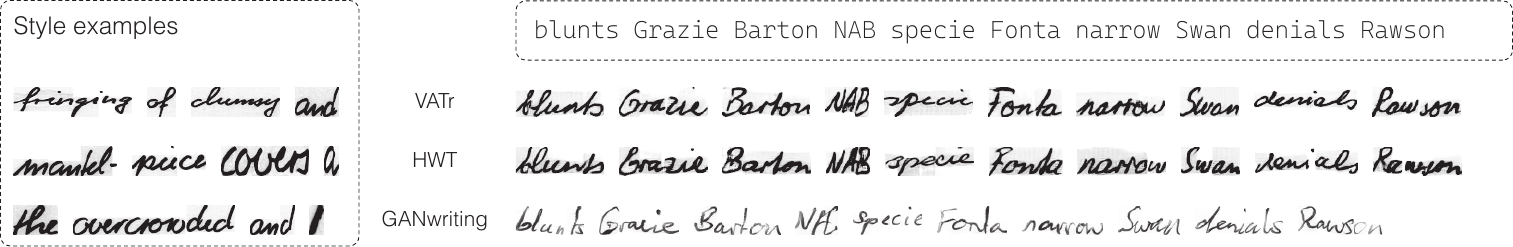} \\
    \vspace{0.5cm}
    \includegraphics[width=\textwidth]{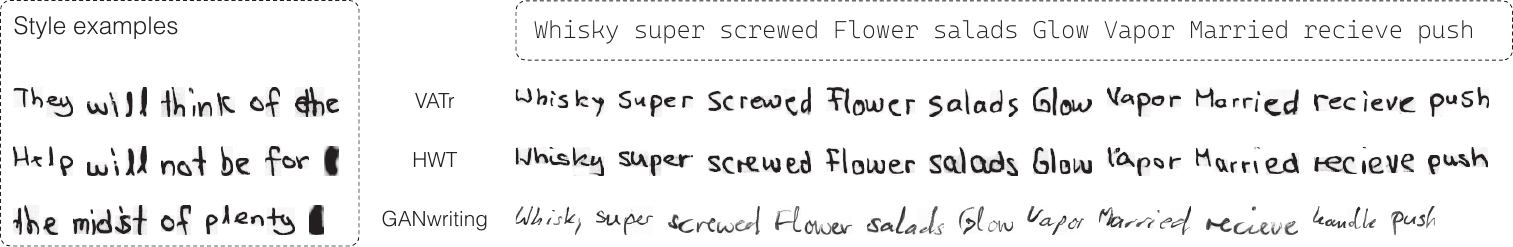} 
    \caption{Exemplar qualitative results of styled text generation for the OOV-U setting on the IAM dataset.}
    \label{fig:iam_oov_u}
\end{figure*}

In Figures~\ref{fig:iam_iv_s}-\ref{fig:iam_oov_u}, we report qualitative examples of images generated with the proposed VATr compared to GANwriting~\cite{kang2020ganwriting} and HWT~\cite{bhunia2021handwriting} in the four styled generation IV-S, IV-U, OOV-S, and OOV-U scenarios, respectively. It can be noticed that the style fidelity in the images produced by VATr does not deteriorate significantly when the style examples are unseen compared with the case in which they have been seen in training. It can be also noticed that VATr consistently disregards the background while reproducing the style, while the HWT competitor reproduces the background artifacts, especially in the unseen style scenarios.

\section{Styled Charsets Generation}\label{sec:charsets}
\begin{figure*}
    \includegraphics[width=\textwidth]{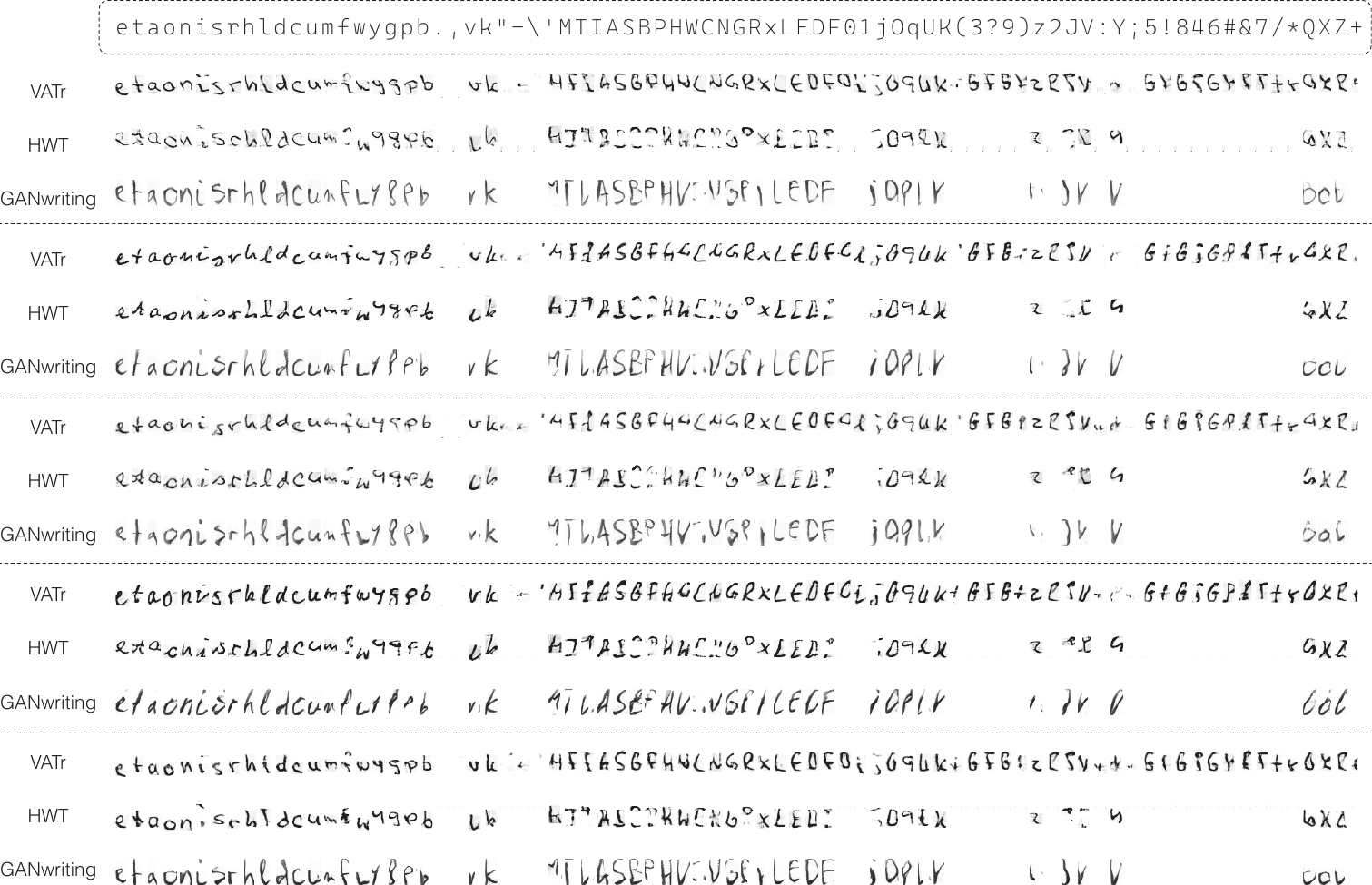}
    \caption{Exemplar IAM charsets generated in different styles sorted by frequency.}
    \label{fig:charsets}
\end{figure*}

In Figure~\ref{fig:charsets}, we report qualitative examples of styled generation of the characters in the IAM dataset, sorted by the frequency in which they appear in the training set. Also in this case, we compare VATr against HWT and GANwriting, which have all been trained on the same data. It has to be noted that, by design, GANwriting cannot generate characters that are not letters. On the other hand, HWT includes also punctuation and digits in its charset. Nonetheless, it can be noticed that VATr generates images for a larger number of characters, including long-tail ones, with which both the competitors struggle.

\begin{table}[]
    \footnotesize
    \centering
    \setlength{\tabcolsep}{.38em}
    \renewcommand{\arraystretch}{0.5}
    \caption{Ablation analysis on the loss terms. The qualitative examples refer to the generation of the word \texttt{that} in two different styles (indicated on top).}
    \label{tab:losses_abl}
    \resizebox{\linewidth}{!}{
    \begin{tabular}{c c c c c c  c c}
    \toprule
    \textbf{$L_{adv}$} & \textbf{$L_{HTR}$} & \textbf{$L_{class}$} & \textbf{$L_{cycle}$} & \textbf{FID} & & \includegraphics[height=0.3cm]{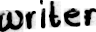} & \includegraphics[height=0.3cm]{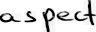}\\
    \midrule
    \cmark &        &        &        & 224.87 & & \includegraphics[height=0.3cm]{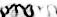} & \includegraphics[height=0.3cm]{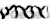} \\ 
    \cmark &        &        & \cmark & 261.23 & & \includegraphics[height=0.3cm]{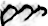} & \includegraphics[height=0.3cm]{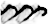} \\ 
    \cmark &        & \cmark &        & 255.11 & & \includegraphics[height=0.3cm]{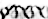} & \includegraphics[height=0.3cm]{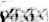} \\ 
    \cmark &        & \cmark & \cmark & 220.83 & & \includegraphics[height=0.3cm]{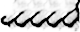} & \includegraphics[height=0.3cm]{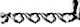} \\ 
    \cmark & \cmark &        &        &  46.61 & & \includegraphics[height=0.3cm]{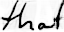} & \includegraphics[height=0.3cm]{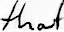} \\ 
    \cmark & \cmark &        & \cmark &  41.68 & & \includegraphics[height=0.3cm]{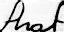} & \includegraphics[height=0.3cm]{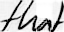} \\ 
    \cmark & \cmark & \cmark &        &  18.70 & & \includegraphics[height=0.3cm]{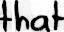} & \includegraphics[height=0.3cm]{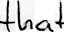} \\ 
    \cmark & \cmark & \cmark & \cmark &  \textbf{17.79} & & \includegraphics[height=0.3cm]{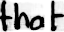} & \includegraphics[height=0.3cm]{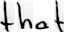} \\ 
    \bottomrule
    \end{tabular}}
\end{table}

\section{Additional Long-tail Character Generation Results}~\label{sec:long-tail}
\begin{figure*}
    \centering
    \includegraphics[width=0.75\textwidth]{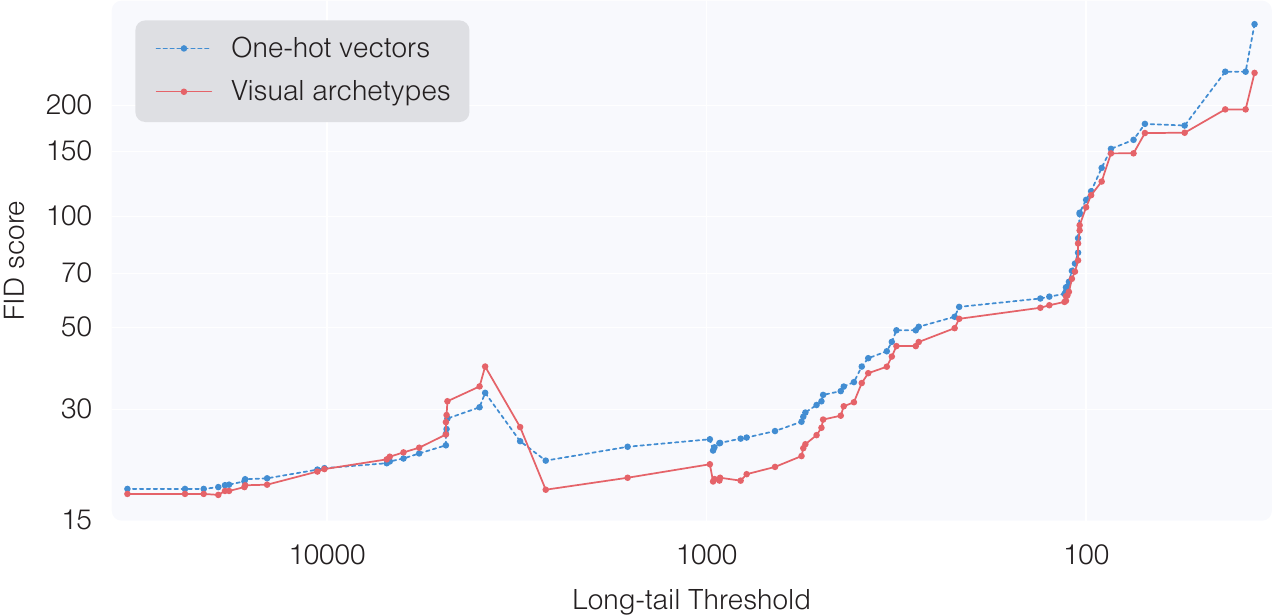} 
    \caption{FID score on words containing long-tail characters with respect to the threshold set to consider a character as rare in the IAM dataset (the x-axis is in logarithmic scale).}
    \label{fig:iam_FID_LT}
\end{figure*}

\begin{figure}
    \centering
    \includegraphics[width=\columnwidth]{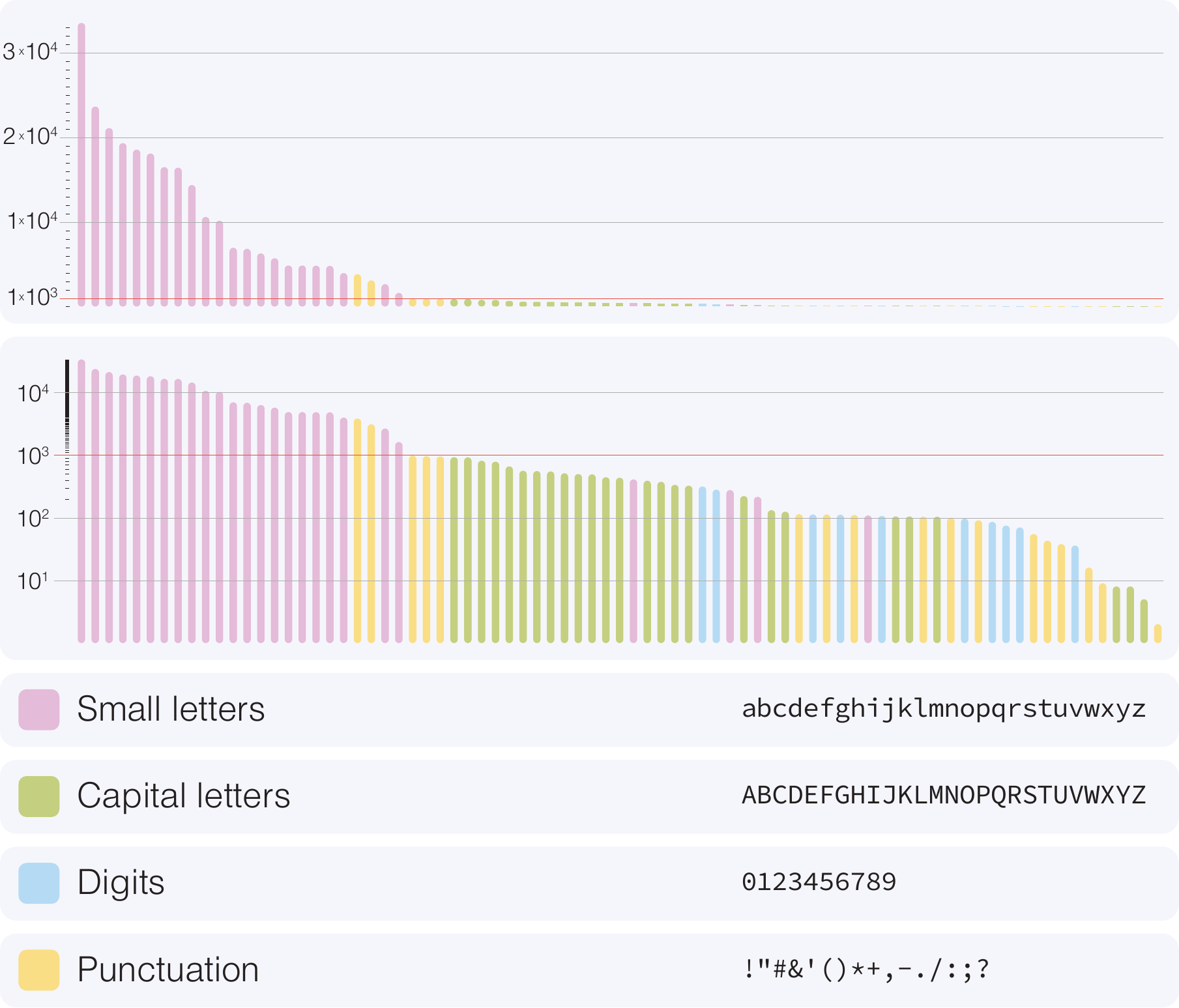} 
    \caption{Distribution and classification of the characters in the training set of the IAM dataset in linear scale (top) and logarithmic scale (bottom).}
    \label{fig:combo_graph}
\end{figure}

We deepen our analysis on the generation of test words containing rare characters in Figure~\ref{fig:iam_FID_LT}, where we compare VATr against HWT in terms of the FID obtained by changing the Long-tail threshold. This is the value of the frequency of appearance in the training set for which a character can be considered as long-tail. Recall that in the experiments in the main paper we set this threshold to 1000, which was determined by observing the character distribution in the IAM training set, reported both in logarithmic and linear scale in Figure~\ref{fig:combo_graph}. 
It can be noticed that the FID obtained by VATr is generally lower than that obtained by HWT, especially at lower threshold values. This demonstrates the robustness of our approach of exploiting dense representations of the characters, compared to resorting to one-hot vectors.
It can be noticed an increase in the FID of both approaches when the threshold is around \num{3500}. Setting the threshold to this value means excluding almost all the words containing only small letters. Afterward, the percentage of capital letters, digits, and punctuation becomes higher than that of small letters in the generated words. Thus, the capacity to faithfully generate also those symbols is more relevant for those words. 

\section{Additional Out-of-Charset results}\label{sec:outofcharset}
\begin{figure*}
    \vspace{0.5cm}
    \includegraphics[width=\textwidth]{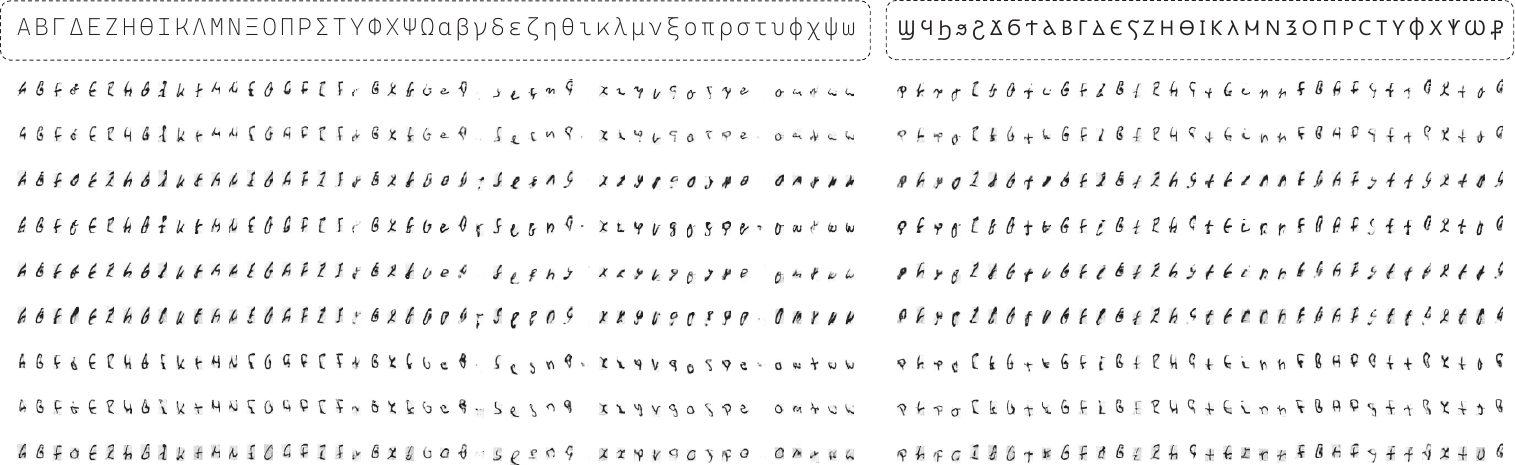} \\
    \caption{Exemplar out-of-charset symbols generated in different styles. The alphabets used are Greek (left) and Coptic (right).}
    \label{fig:ooc}
\end{figure*}

In Figure~\ref{fig:ooc}, we report examples of the generation of out-of-charset characters from non-Latin alphabets (Greek and Coptic) with our proposed approach. Despite this setting being beyond the scope of our work, it can be noticed that VATr is able to reproduce symbols that are not present in its training set, especially in the case these are geometrically similar to in-charset symbols. In fact, when generating Greek letters that are close to Latin ones starting from their respective visual archetypes, VATr can exploit learned geometric regularities. On the other hand, it struggles to generate characters from the Copto alphabet, whose visual archetypes are very different from those of Latin letters.

\section{Styled HTG for HTR}\label{sec:htr}
One of the main applications of styled HTG is providing training data for HTR models to be applied to writer-specific manuscripts. To assess the potential benefits of this strategy when using VATr to improve an HTR model in this setting, we consider  the interesting case of two low-resource single-author HTR datasets (Saint Gall~\cite{fischer2011transcription} and Washington~\cite{fischer2012lexicon}\footnote{\url{https://fki.tic.heia-fr.ch/databases/iam-historical-document-database}}) and generate synthetic training lines with VATr. We then use these as additional samples to train a SotA HTR model~\cite{cojocaru2020watch,cascianelli2022boosting} and compare the transcription results against a baseline not exploiting synthetic data and the baseline exploiting synthetic data generated with the one-hot vectors-based HWT~\cite{bhunia2021handwriting}. These results are reported in Table~\ref{tab:htr}, expressed in terms of Character Error Rate (CER) and Word Error Rate (WER). It can be observed that training on VATr-generated images reduces the errors \wrt both the compared approaches, especially in terms of WER. This suggests that the misrecognized characters are more concentrated in single words.

\begin{table}[t]
\footnotesize
\centering
\setlength{\tabcolsep}{.32em}
\caption{Performace comparison of an HTR model trained on real data only and on a combination of real and generated styled text images (obtained with our approach and the SotA HWT approach) on two writer-specific datasets.}
\label{tab:htr}
\resizebox{0.85\linewidth}{!}{
\begin{tabular}{l c cc c cc}
\toprule
 && \multicolumn{2}{c}{Saint Gall} && \multicolumn{2}{c}{Washington}\\
 \cmidrule{3-4} \cmidrule{6-7}
 && CER & WER && CER & WER \\
\midrule
Real && \textbf{4.5} & 32.5 && 3.4 & 15.9 \\
\midrule
HWT + Real && 4.6 & 31.2 && 3.7 & 16.5 \\
VATr + Real && \textbf{4.5} & \textbf{30.9} && \textbf{3.0} & \textbf{13.1} \\

\bottomrule
\end{tabular}
}\vspace{0.45cm}
\end{table}

{\small
\balance
\bibliographystyle{ieee_fullname}
\bibliography{supplementary_main}
}